\documentclass[journal,twoside,web]{ieeecolor}
\usepackage{tmi}
\usepackage{cite}
\usepackage{amsmath,amssymb,amsfonts}
\usepackage{algorithmic}
\usepackage{graphicx}
\usepackage{textcomp}
\usepackage{bbm}
\usepackage{multirow}
\usepackage[colorlinks,linkcolor=black]{hyperref}
\def\BibTeX{{\rm B\kern-.05em{\sc i\kern-.025em b}\kern-.08em
    T\kern-.1667em\lower.7ex\hbox{E}\kern-.125emX}}
\begin{document}
\title{Graph Flow: Cross-layer Graph Flow Distillation for Dual Efficient Medical Image Segmentation}
\author{Wenxuan Zou, Muyi Sun
\vspace{-10mm}
\thanks{This work was done when Wenxuan Zou was an intern in CASIA under the guidance of Muyi Sun. This work is the continuation of Muyi Sun's research on Medical Image Analysis. Muyi Sun bears the primary responsibility.(Corresponding authors: Muyi Sun)}
%This work is supported in part by NSFC under Grant BS123456. 
\thanks{This work is supported by China Postdoctoral Science Foundation Under Grant 2022M713362.
}
\thanks{Wenxuan Zou is with the School of AI/Automation, Beijing University of Posts and Telecommunications, Beijing, China, 100876, (e-mails: zouwenxuan@bupt.edu.cn).}
\thanks{Muyi Sun is with the CRIPAC, NLPR, Institute of Automation, Chinese Academy of Sciences, Beijing, China, 100190, (e-mail: muyi.sun@cripac.ia.ac.cn).}
%\thanks{Caifeng Shan is with the Shandong University of Science and Technology, Qingdao, China, 266590 and also with the Artificial Intelligence Research, Chinese Academy of Sciences (CAS-AIR), Beijing, China, 100190, (e-mail: caifeng.shan@gmail.com).}
}

\maketitle

%%%%ABSTRACT需要改！！！！！！！！！！！！
\begin{abstract}
With the development of deep convolutional neural networks, medical image segmentation has achieved a series of breakthroughs in recent years. However, the high-performance convolutional neural networks always mean numerous parameters and high computation costs, which will hinder the applications in clinical scenarios.
Meanwhile, the scarceness of large-scale annotated medical image datasets further impedes the application of high-performance networks.
To tackle these problems, we propose Graph Flow, a comprehensive knowledge distillation framework, for both network-efficiency and annotation-efficiency medical image segmentation. 
%Specifically, our core Graph Flow Distillation, transfer the cross-layer graph flow knowledge from a well-trained cumbersome teacher network to a non-trained compact student network.  
Specifically, our core Graph Flow Distillation transfer the essence of cross-layer variations from a well-trained cumbersome teacher network to a non-trained compact student network.  
In addition, an unsupervised Paraphraser Module is integrated to purify the knowledge of the teacher network, which is also beneficial for the stabilization of training procedure. 
%Furthermore, we build a unified distillation framework by integrating the adversarial distillation and the vanilla logits distillation, which can further promote the final performance respectively. 
Furthermore, we build a unified distillation framework by integrating the adversarial distillation and the vanilla logits distillation, which can further refine the final predictions of the compact network. 
%Extensive experiments conducted on Gastric Cancer Segmentation Dataset and Synapse Multi-organ Segmentation Dataset demonstrate the prominent ability of our method which achieves state-of-the-art performance on these different-modality and multi-category medical image datasets.
With different teacher networks (conventional convolutional architecture or prevalent transformer architecture) and student networks, we conduct extensive experiments on four medical image datasets with different modalities (Gastric Cancer, Synapse, BUSI, and CVC-ClinicDB).We demonstrate the prominent ability of our method which achieves competitive performance on these datasets.
Moreover, we demonstrate the effectiveness of our Graph Flow through a novel semi-supervised paradigm for dual efficient medical image segmentation.
Our code will be available at \href{https://github.com/WrinkleXuan/Graph-FLow}{\textit{Graph Flow}}.
%The code will be available at https://github.com/WrinkleXuan/Graph-FLow.
\end{abstract}

\vspace{-3mm}

\section{Introduction}
\label{sec:introduction}
\IEEEPARstart{M}{edical} image segmentation (MIS), a challenging computer vision task, aims to automatically assign pixel-wise labels for the specific regions in medical images, such as organs, cells, lesion areas, as precisely as possible. 
In the field of computer-aided diagnosis and therapy, medical image segmentation plays a crucial role, which can efficiently alleviate the burden for clinicians to identify pathological or anatomical regions.
In the past decade, the employment of Convolutional Neural Networks (CNNs) has become a milestone for medical image segmentation. 
Various elaborate CNNs\cite{ronneberger2015u, milletari2016v, zhou2019unet++,isensee2021nnu, li2020accurate} exhibit superior performances on medical images of different modalities.
\begin{figure}[t]
\begin{center}
\includegraphics[width=\linewidth]{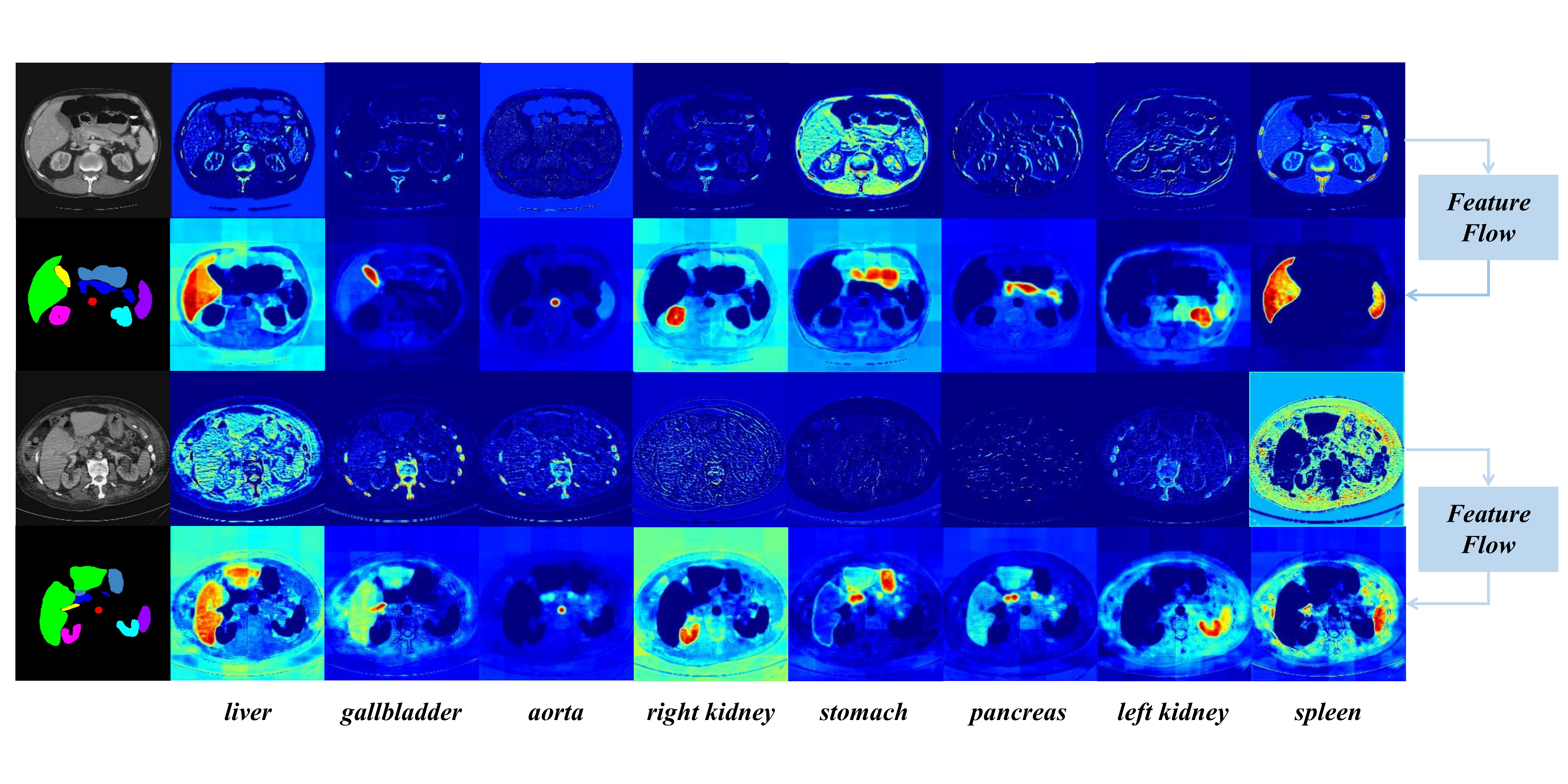}
\end{center}
%\vspace{-3mm}
    \caption{The illustrations of feature changes (flow) from shallow to deep layers. 
    The first and third rows visualize the sampled feature maps of shallow layers, and the second and fourth rows visualize the corresponding sampled feature maps of depth layers.
    The changes between the first two rows visualize the variations of the same-channel activations from shallow layer to deep layer, as well as the last two rows.
    In the first column, CT scans and organ annotations from two different patients are listed from top to bottom. 
    We could observe that the attention of the segmentation network changes from a shallow layer to a deep layer, and meanwhile the feature channels show more distinct class-aware salience regions in the deeper layer.
    Therefore, we regard these changes as a type of distillation knowledge and design a Graph Flow Distillation to model these changes for efficient MIS.
    }
    %\vspace{-3mm}
\label{fig:heatmaps}
\end{figure}
%However, such tailored CNNs always consist of sophisticated network architectures with millions of parameters and vast computational operations. 
Such tailored CNNs always consist of sophisticated network architectures with millions of parameters and vast computational operations. 
However, these networks are degraded or even incompetent if employed in the scenes with limited
computational resources such as in the rural clinics of developing countries, or in the
real-time diagnosis scenario of large hospitals with tens of thousands of patients daily.
These situations are further aggravated when processing the medical images with huge resolution and deep volumes (e.g. pathological Whole-Slide-Image (WSI)\cite{takahama2019multi,tokunaga2019adaptive} and Computed Tomography (CT)\cite{cciccek20163d,lei2020medical}), or operating at bedsides on mobile devices (e.g. ultrasound image\cite{xue2021global} or colonoscopy image\cite{kim2021uacanet}).
%Meanwhile, the medical images are generally involved huge resolution or deep volumes.
%Concretely, the resolution of a pathological Whole-Slide-Image (WSI) (e.g. $10^5\times10^5$ pixels\cite{takahama2019multi}) is over 100 times a natural image\cite{tokunaga2019adaptive}. 
%The Computed Tomography (CT) and Magnetic Resonance Imaging (MRI) are volumetric data which consist of hundreds of slices\cite{cciccek20163d,lei2020medical}. 
In addition, the annotation of medical images is expensive due to the requirements of experienced experts and abundant time.
Thus, large-scale annotated medical image datasets are always scarce. 
%These above characteristics of CNNs and medical images result in a high-performance network with less computation and storage costs is unavailable in resource-limited and label-limited medical application scenes. 
The above characteristics of the CNNs and medical images result in that the high performance networks with less computation and storage costs are unavailable in resource-limited and label-limited medical application scenes.
%Therefore, the trade-off between the accuracy and cost of CNNs needs to be considered.
Therefore, the trade-off between the performance and cost of CNNs needs to be considered. 
At the same time, the massive and available unlabeled medical data should also be taken into account as much as possible, due to the lack of annotated medical images in real applications. 

%To tackle these dilemmas, most researchers engage in devising lightweight networks in the medical imaging community, due to the small demand for resources and annotated medical images.
%\textcolor{blue}{To tackle these dilemmas, most researchers engage in devising lightweight networks in the medical imaging community, due to well-trained cumbersome CNNs without over-fitting always need large-scale and fine-grained annotated medical image datasets and consume more resources.}
To tackle the resource-limited dilemma, some researchers engage in devising lightweight networks in the medical imaging community, since lightweight networks consume less computational resource and inference time than cumbersome networks.
Zhou \emph{et al.}\cite{zhou2020lightweight} devise a double attention network with a pre-trained MobileNetV2 encoder for real-time stent segmentation in intraoperative X-ray ﬂuoroscopy.
Lei \emph{et al.}\cite{lei2020lightweight} propose a lightweight V-Net which consists of inverted residual bottleneck blocks based on depthwise separable convolution for volumetric liver tumor segmentation.
Zhang \emph{et al.}\cite{zhang2021lpaqr} present a lightweight vertebra segmentation network, termed LPAQR-Net, which utilizes condensed feature channels and early down-sampling strategy to reduce the number of parameters and shorten the inference time.
%However, these lightweight networks are not always able to satisfy the requirement of performance in real medical scenarios, as these compact architectures could not extract more representative pathological or organic features than cumbersome networks. 
However, when cumbersome networks are impracticable in some harsh medical situations, such as in some poverty-stricken rural hospitals and real-time diagnosis requirements for developed hospitals, these lightweight networks are not always able to satisfy the requirement of performance, as these compact architectures could not extract more representative pathological or organic features than cumbersome networks.

Recently, knowledge distillation, a straightforward and effective model compression technique, has drawn more and more attention in computer vision. 
Knowledge distillation is designed to distill and transfer the important knowledge hidden in a pre-trained cumbersome network (teacher network) to a non-trained compact network (student network)\cite{gou2021knowledge}.
Meanwhile, knowledge distillation can promote the existing compact networks to extract more informative features than themselves trained from scratch, without redesigning delicate compact architectures. 
In the field of computer vision, the initial idea of knowledge distillation is to mimic the softened logit distributions between the teacher network and the student network\cite{hinton2015distilling}. 
Subsequently, researchers introduce knowledge distillation into the intermediate features of neural networks by direct approximating the features of student networks \cite{romero2014fitnets}. 
Nevertheless, it is difficult for the student network to mimic the intermediate representations of the teacher network with different network architecture.
To this end, more studies based on relation knowledge (e.g. class-wise and layer-wise relation) are proposed \cite{tung2019similarity,peng2019correlation,liu2020structured,wang2020intra,yim2017gift} recently. 
%Concretely, the class-wise relation distillation methods\cite{liu2020structured,wang2020intra} are proposed based on pixel-wise similarity in the field of image segmentation. 
As for class-wise relation distillation\cite{liu2020structured,wang2020intra}, these methods model the pixel-level correlations between classes but lack holistic class-wise structural information, which might damage the exact segmentation of crucial areas. 
%In layer-wise relation distillation, the Flow of Solution Procedure (FSP)\cite{yim2017gift} is defined as a type of high-level distilled knowledge which can prompt the performance of the student network. 
%Nevertheless, the direct computation of the FSP matrix would neglect the inter-class variations from low-level representations to high-level representations and include various redundant or even harmful knowledge. 
%On the other hand, the layer-wise relation distillation \cite{yim2017gift} would transfer various redundant or even harmful knowledge. 
On the other hand, the layer-wise relation distillation \cite{yim2017gift} would transfer various redundant knowledge.

Though the knowledge distillation has developed deeply in computer vision,  only a few researchers try to explore knowledge distillation in the medical imaging community.
%Wen \textit{et al.}\cite{wen2021towards} propose a boundary-guided distillation method for organ segmentation. 
%Qin \textit{et al.}\cite{qin2021efficient} design a region affinity distillation method for efficient liver tumor segmentation. 
%Tran \textit{et al.}\cite{tran2022light} propose a new Light-weight Deformable Registration network that introduce a new adversarial learning with distilling knowledge.  
%However, their methods mostly are built on some basic concepts of knowledge distillation but do not notice the variations of pathological or organic representations from different layers.
Wen \textit{et al.}\cite{wen2021towards} propose a boundary-guided distillation method for organ segmentation. 
Qin \textit{et al.}\cite{qin2021efficient} design a region affinity distillation method for efficient liver tumor segmentation. 
Tran \textit{et al.}\cite{tran2022light} propose a new Light-weight Deformable Registration network that introduces a new adversarial learning with distilling knowledge.  
Since their methods are primary explorations for utilizing knowledge distillation in medical imaging analysis, they do not notice the variations of pathological or organic representations from different layers.
%文章后面就用inter-class吧 
%这个地方的总结 把CV中的KD和医学的KD 的缺陷整体描述 然后引出下面我们的工作。

In this paper, considering the properties of medical datasets, segmentation network architectures, and the mentioned drawbacks of knowledge distillation, we propose a novel knowledge distillation framework, \textbf{Graph Flow}, to promote the trade-off between the performance and efficiency in efficient medical image segmentation. 
It worth emphasizing that the mentioned `efficiency' involves two meanings: \textbf{network efficiency and annotation efficiency.}
In other words, our method could not only compress the complexity of high-performance networks to satisfy the requirement of resource-limited medical scenarios, but also exploit the massive unlabeled medical images to alleviate the lack of annotations. 

%跨层之间的变化是展现了如何解决问题的过程，这比教学生答案更重要
We visualize the motivation of our method on the Synapse Multi-organ Segmentation Dataset, as shown in Fig.~\ref{fig:heatmaps}.
The feature flow of cross-layers exhibits the procedures of solution questions of CNNs, which is more critical than teaching students to remember answers.
However, the flow of cross-layers contains redundant or useless knowledge, which may confuse the students to comprehend such procedures well.
As illustrated in Fig.~\ref{fig:heatmaps}, the essence of cross-layer flow is the variation of structural semantic-aware knowledge, while the semantic-aware knowledge can be represented well based on an effective graph.
Specifically, we firstly filter out the patch-wise salience regions based on the maximum activation area.
Next, we suppress the activations of regions except for these salience regions along the axis of feature channels.
After that, we propose our \textbf{salience graph} to encode structural semantic-aware knowledge without any other useless knowledge beyond these channel-wise and patch-wise salience regions.
At last, we construct our \textbf{variation graph} to measure the flow of salience graphs between different layers.
Therefore, our method, called \textbf{Graph Flow Distillation}, distills the variation graph from teacher networks to student networks. 
In addition, we utilize a Paraphraser Module\cite{kim2018paraphrasing} to refine the representations of teacher network. 
Moreover, we integrate the adversarial training strategy\cite{xu2017training} (Adversarial Distillation) and the vanilla knowledge distillation\cite{hinton2015distilling} (Logits Distillation) into our Graph Flow to establish a full-stage knowledge distillation framework, which are both beneficial to the final performance.

On account of the intrinsic characteristics of knowledge distillation, our method could also transfer knowledge from a cumbersome teacher (pre-trained with the limited labeled data) without the supervision of annotation.
Thus, our Graph Flow could be further utilized to conquer the scarceness of annotations for efficient medical image segmentation in a semi-supervised manner. 

Our main contributions are as follows:

$\bullet$ We propose Graph Flow Distillation, a novel knowledge distillation method, to transfer the flow of cross-layer salience graphs from a teacher network to a student network.

%$\bullet$ We design a Paraphraser Module to purify the representations of the teacher network, which stabilizes the network training procedure.%体现稳定

%$\bullet$ We \textcolor{blue}{utilize} a Paraphraser Module to purify the representations of the teacher network, which stabilizes the network training procedure.%体现稳定
%$\bullet$ We propose a unified distillation framework, dubbed Graph Flow, to incorporate Graph Flow Distillation, Adversarial Distillation, and Logits Distillation, which facilitates efficient knowledge transfer.

$\bullet$ We propose a unified distillation framework, dubbed Graph Flow, to incorporate Graph Flow Distillation, Adversarial Distillation, and Logits Distillation, and further provide a new paradigm for semi-supervised efficient medical image segmentation based on our Graph Flow with a tweaking on the objective functions.

%$\bullet$ We design a Paraphraser Module to purify the representations of the teacher network, which stabilizes the network training procedure.%体现稳定

$\bullet$ We conduct extensive experiments on the different-modality and multi-category medical image segmentation datasets. Extensive quantitative and qualitative analyses demonstrate the state-of-the-art ability of our Graph Flow. 

This work has several noteworthy extensions compared with our preliminary work\cite{zou2021coco}: 
$1)$ We distill the flow of cross-layer salience graphs by variation graph instead of the variations of channel-mixed spatial similarity between different layers.
$2)$ We propose a unified framework which integrates previous knowledge distillation methods for efficient knowledge transfer.
$3)$ We verify our method on more medical image segmentation datasets with different modalities and multiple categories.
$4)$ We provide a new paradigm for semi-supervised efficient medical image segmentation.

\section{Related Work}
\subsection{Medical Image Segmentation}
Medical image segmentation engages in automatic labeling organic or pathological structures in medical images. 
In recent years, medical image segmentation has achieved impressive advances with the development of deep learning.
U-Net\cite{ronneberger2015u}, consisting of a symmetric architecture and skip connections, has served as a backbone network in medical image segmentation. 
Zhou \emph{et al.}\cite{zhou2019unet++} design UNet++ with a new nested architecture and dense-skip connections for improving the details of medical image segmentation. 
Milletari \emph{et al.}\cite{milletari2016v} present V-net which utilizes 3D convolution and residual connections for volumetric medical data.
%Li \emph{et al.}\cite{li2018h} proposed a hybrid densely-connected UNet, called H-DenseUNet, for liver tumor segmentation. 
Isensee \emph{et al.}\cite{isensee2021nnu} propose a novel pipeline from self-configuring preprocessing and post-processing of dataset to U-shape network architecture.
Though the above networks achieve the state-of-the-art the performance in different modalities of medical images, they are not applicable in some harsh medical scenes due to the incredible high storage, computational, and annotation cost.
Therefore, some researchers intend to devise lightweight network architectures. 
Zhang \emph{et al.}\cite{zhang2019light} devise a lightweight convolutional neural network with 2D and 3D convolution jointly. 
Guo \emph{et al.}\cite{guo2021sa} propose a lightweight Spatial Attention U-net for retinal vessel segmentation. 
Nevertheless, such compact networks result in performance degradation with large-scale networks. 
It is urgent to propose an efficient method which can circumvent the trade-off between the performance and efficiency in medical image segmentation.

\subsection{Knowledge Distillation}

The knowledge distillation can be briefly defined as the training of a minor student network with the beneficial knowledge transferred from a cumbersome teacher network. 
Hinton \emph{et al.}\cite{hinton2015distilling} firstly propose the notation of knowledge distillation and demonstrate that the soft targets of the teacher network can prompt the performance of the student network. 
Next, Romero \emph{et al.}\cite{romero2014fitnets} propose FitNets using intermediate hidden layers of a teacher with a guided layer. 
However, such straightforward feature-based distillation methods are not sufficient for generalization.
%Therefore, various relation-based distillation methods are explored.
%Peng \emph{et al.}\cite{peng2019correlation} propose a new framework called Correlation Congruence Knowledge Distillation, which distills the correlation between different samples. 
%Chen \emph{et al.}\cite{chen2020improving} transfer category-level structural knowledge between different instances. 
In the field of image segmentation, most knowledge distillation studies focus on relation-based knowledge.
Liu \emph{et al.}\cite{liu2020structured} utilize pair-wise distillation and holistic distillation to guide the learning of the student. 
Shu \emph{et al.}\cite{shu2021channel} mimic the Kullback–Leibler divergence of the channel-wise probability maps between the teacher and the student. 
Hou \emph{et al.}\cite{hou2020inter} design an inter-region graph based on the similarity for road marking segmentation. 
Meanwhile, other previous works also explore the graph-based methods\cite{liu2019knowledge,zhou2021distilling, lee2019graph}.
Liu  \emph{et al.}\cite{liu2019knowledge} design a relationship graph (IRG) between the images (instances) in the same batch to measure the intra-class variations and the inter-class differences based on instance-level features.
Zhou \emph{et al.}\cite{zhou2021distilling} construct a unified graph-based embedding, called attributed graph, by aggregating individual knowledge from relational neighborhood images. 
Lee \emph{et al.}\cite{lee2019graph} propose Multi-Head Graph Distillation (MHGD) for transferring the intra-data relation knowledge between samples in the mini-batch.
However, these methods are all designed based on instance-level features for image classification;
compared to our method, they can not transfer the pixel-level and semantic-aware knowledge.
Besides, only a few researchers introduce knowledge distillation for efficient medical image segmentation.
Wen \textit{et al.}\cite{wen2021towards} propose a boundary-guided knowledge distillation method which assists the student network to align organ boundary features in teacher network. 
However, the effectiveness of this method depends on the accuracy of boundary extraction.
Qin \textit{et al.}\cite{qin2021efficient} design a region affinity distillation method, and integrate the importance map for efficient liver tumor segmentation, while this method perform unsatisfactory due to the mismatching of auxiliary masks and feature maps.
Moverover, these methods mostly concentrate on distilling the knowledge from certain layers, while do not notice the variations of pathological or organic semantic-aware representations from different layers. 
Inspired by the mentioned knowledge distillation methods and the intrinsic property of medical images, we propose Graph Flow for dual efficient medical image segmentation.

\subsection{Semi-supervised Medical Image Segmentation}
The achieved remarkable successes of deep learning in medical image segmentation are widely dependent on massive high-quality annotated datasets. 
However, such datasets are usually unavailable since medical datasets are expensive to annotate\cite{tajbakhsh2020embracing}. 
Semi-supervised learning can relieve the scarcity of dataset annotation by exploring unlabeled data. 
Yu \emph{et al.}\cite{yu2019uncertainty} propose a semi-supervised medical segmentation framework based on uncertainty-awareness in which an uncertainty consistent loss can prompt the accuracy of segmentation. 
%Luo \emph{et al.}\cite{luo2021semi} build a dual-task regularization method which enforces a task-level consistency between a pixel-wise segmentation map and a geometry-aware target representation for 3D gadolinium-enhanced MRI analysis. 
%With the geometric shape constraints hidden in unlabeled medical data, Li \emph{et al.}\cite{li2020shape} presented a multi-task semi-supervised network which employs adversarial loss between labeled and unlabeled data for 3D medical image segmentation. 
Yang \emph{et al.} \cite{yang2021medical} design a novel hybrid loss function for semi-supervised medical instrument segmentation, consisting of a uncertainty and contextual constraints.
Nevertheless, current semi-supervised medical image segmentation methods do not take account of the limitation of computational resources and memory storage in real scenes, while our Graph Flow can solve the lack of annotated datasets and the limitation of computational resources and memory storage simultaneously.

%2022.3.15 dual-efficient
%Our semi-supervised strategy is beyond our lightweight distillation framework.
Therefore, our semi-supervised strategy is embedded in our distillation method for dual efficient medical image segmentation, which is different with other semi-supervised methods.
\begin{figure*}[t]
\begin{center}
\includegraphics[width=0.9\linewidth]{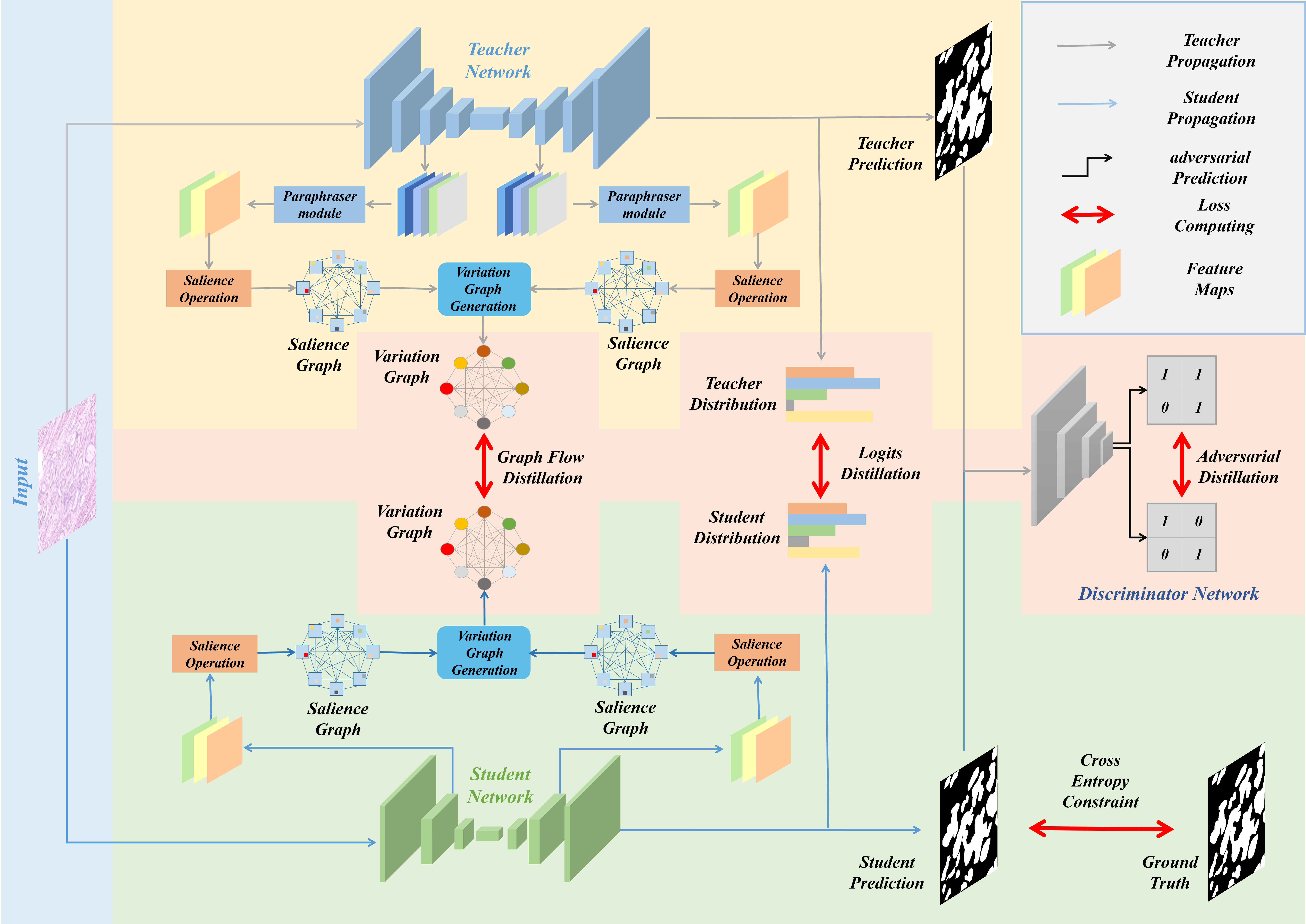}
\end{center}
   \vspace{-5mm}
   \caption{The overall framework of our Graph Flow. As a comprehensive knowledge distillation framework, it integrates three different strategies: Graph Flow Distillation, Adversarial Distillation, and Logits Distillation. In Graph Flow, Paraphraser Module is committed to stable and refine the knowledge transfer. In this framework, the network with yellow background represents the teacher network, the network with green background represents the student network, and the light pink background area represents the distillation operations. }
    \vspace{-5mm}
\label{fig:framework}
\end{figure*}

\vspace{-5mm}
\section{Method}
The purpose of our Graph Flow is to improve the performance of the non-trained compact target student network $S$ under the extra supervision of the knowledge transferred from the cumbersome teacher network $T$ in the medical image segmentation task.
The pipeline of our Graph Flow is illustrated in Fig.~\ref{fig:framework}, which consists of three different distillation methods including Graph Flow Distillation, Adversarial Distillation, and Logits Distillation. 
Our Graph Flow Distillation takes into account cross-layer knowledge flow, as shown in Fig.~\ref{fig:heatmaps}, which is neglected by previous knowledge distillation methods in medical image segmentation.
Meanwhile, we also introduce Adversarial Distillation and Logits Distillation that constrain knowledge consistency between the outputs of $T$ and $S$ combined with distilling knowledge from intermediate layers to fulfill the full-stage distillation.
Moreover, our method can simultaneously solve resource-limited and label-limited dilemmas of medical image segmentation in clinical diagnosis. 
Therefore, our Graph Flow is a comprehensive framework for dual efficient medical image segmentation.

To begin with, given a medical image $X$ as input, the teacher network $T$ which is pre-trained by ourselves produces features maps $F_{t_{i}}$ from $i_{th}$ layer, and outputs $Y_{t}$. 
Next, we employ the mentioned distillation methods to transfer different knowledge hidden in intermediate layers and outputs of $T$ for training the student network $S$ with the same input. 
Concretely, our Graph Flow Distillation constructs and matches variation graphs between $S$ and $T$ for distilling the cross-layer knowledge flow.
In addition, Adversarial Distillation utilizes a discriminator network $D$ to align high-order knowledge between $Y_{s}$ and $Y_{t}$. 
At the same time, Logits Distillation is employed to mimic the logits distributions between $T$ and $S$. 
At last, the student network $S$ trained by our Graph Flow can achieve competitive performance compared with the teacher network $T$ without any extra memory and parameters.

\subsection{Graph Flow Distillation}
As illustrated in Fig.~\ref{fig:heatmaps}, the attention of medical image segmentation network greatly changes between different layers. 
Therefore, such knowledge hidden in the flow of cross-layers of teacher network can teach student network to extract more critical features for accurate organic or pathological segmentation.
However, previous layer-wise relation distillation\cite{yim2017gift} is inappropriate for medical image segmentation because it simply calculates the relation between different layers so that does not focus on the variations of semantic-aware knowledge. 
%It is damaged to exact segmentation of similar organs or areas (e.g. left and right kidney). 
Inspired by this phenomenon, we propose our salience graph, which can focus more on structural semantic-aware knowledge (class-wise spatial information and inter-class correlation) of an intermediate layer. 
Our salience graph can distinguish ambiguous semantic categories well, which is beneficial for capturing class-wise spatial information and inter-class correlation.
Furthermore, we devise a variation graph, which can simultaneously represent the flow of class-wise spatial information and inter-class correlation embedded by salience graph from cross-layers.

\begin{figure}[t]
\begin{center}
\includegraphics[width=0.9\linewidth]{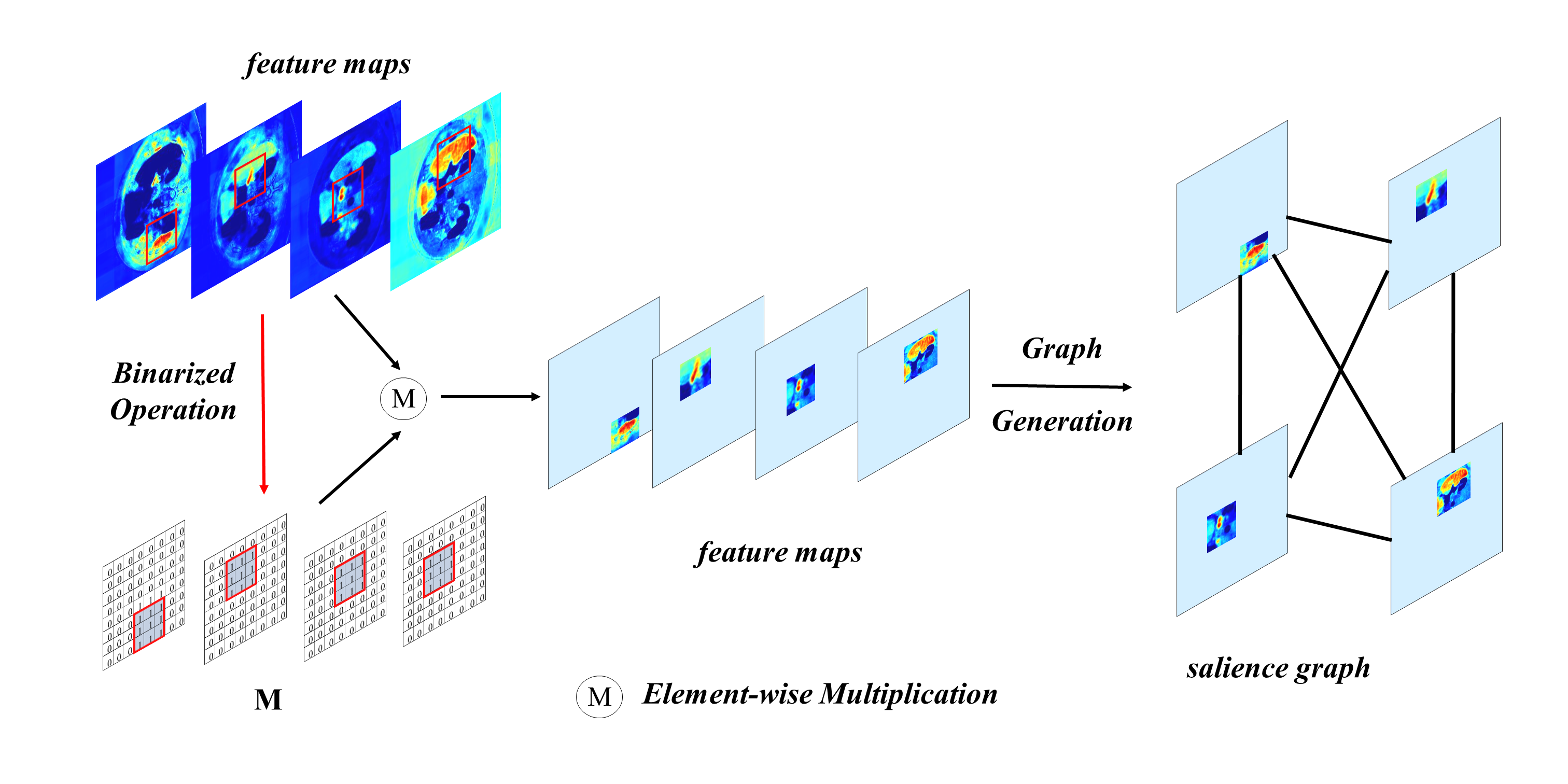}
\end{center}
   \vspace{0mm}
   \caption{The illustration of Salience Operation. The feature maps are suppressed by the binary mask $M$, except for the patch-wise salience regions. Given such feature maps, the salience graph is generated to encode holistic class-wise spatial information and inter-class correlation.}
  \vspace{-6mm}
   \label{fig:salience graph}
\end{figure}

\textbf{\emph{1) Salience Graph:}} For feature maps $F_{(\cdot)_i}$ with the size of $C\times H\times W$ ($C$ is the number of channels, $H\times W$ is the spatial size), we propose salience graph denoted as $SG_{(\cdot)_i}$ ($(\cdot)$ refers to $s$ or $t$) to represent structural semantic-aware knowledge hidden in channel-wise salience features. The construction of $SG_{(\cdot)_i}$ is depicted in Fig.~\ref{fig:salience graph}, which could be formulated as:
\begin{align}
    SG_{(\cdot )_i}=\left ( \Gamma _{(\cdot)_i}, \Theta_{(\cdot)_i}\right),
\end{align}
\begin{align}
     \Gamma_{(\cdot)_i}=MF_{(\cdot)_i},
     \label{eqution2}
\end{align}
\begin{align}
    \Theta_{(\cdot)_i}\left ( c,k \right )= \Phi(\Gamma_{(\cdot)_i}(c),\Gamma_{(\cdot)_i}(k)) (c,k=1,2,\cdots,C)
\end{align}
where $\Gamma_{(\cdot)_i}$ and $\Theta_{(\cdot)_i}$ are the vertex and edge set of $SG_{(\cdot )_i}$ respectively.  
$\Phi(\cdot,\cdot)$ is a function for measuring the correlation between different channel-wise salience regions. 
Each vertex in the $SG_{(\cdot)_i}$ represents one of the channels in the $F_{(\cdot)_i}$. 
As shown in Equation \ref{eqution2}, we utilize $M=[M_{1},M_{2},\cdots,M_{c},\cdots,M_{C}]$ to filter the redundant activations except for the patch $P=[P_{1},P_{2},\cdots,P_{c},\cdots,P_{C}]$ centered on the pixel with the largest activation in every channel. The $M$ is a binary mask with the same size as $F_{(\cdot)_i}$. 
The value of each pixel of $M_{c}$ is computed by $M_{c}(p)=\mathbbm{1}(p\in P_{c}) (p=1,2,\cdots,H\times W)$, where $\mathbbm{1}(\cdot)$ is an indicator function, and $p$ is a pixel of $c_{th}$ channel of $F_{(\cdot)_i}$. 
The different sizes of $P$ indicate that salience graph $SG$ has different sizes of respective field about class-wise knowledge. 
After that, each vertex can encode a specific category with holistic spatial information. 
Each edge of $SG_{(\cdot)_i}$, $\Theta_{(\cdot)_i}\left ( c,k \right )$, are computed by a score function $\Phi(\cdot,\cdot)$, which can encode inter-class correlation and does not destroy holistic structure of the category. We simply define the score function $\Phi(\cdot,\cdot)$ as Euclidean distance in this paper.

\textbf{\emph{2) Variation Graph:}} Assume that we have a pair salience graphs, $SG_{(\cdot)_i}$ and $SG_{(\cdot)_{i^{'}}}$. We design a novel variation graph to measure the flow of salience graphs from $i_{th}$ layer to $i^{'}_{th}$ layer, in which the $i^{'}_{th}$ layer is a corresponding layer of $i_{th}$ layer with same size $H\times W\times C$. The variation graph, termed $VG_{(\cdot)_i}$, can be presented as:
\begin{align}
    VG_{(\cdot)_i}=(V_{(\cdot)_i}, E_{(\cdot)_i}),
\end{align}
\begin{align}
    V_{(\cdot)_i}(c)= \left\| \Gamma_{(\cdot)_i}(c)-\Gamma_{(\cdot)_{i^{'}}}(c)\right\|_{2}^{2}   (c=1,2,\cdots,C),
    \label{equation5}
\end{align}
\begin{align}
    E_{(\cdot)_i}(c,k)=\left\| \Theta_{(\cdot)_i}(c,k)-\Theta_{(\cdot)_{i^{'}}}(c,k)\right\|_{2}^{2} (c,k=1,2,\cdots,C)
\end{align}
where $V_{(\cdot)_i}$ and $E_{(\cdot)_i}$ are the vertex and edge set of $VG_{(\cdot)_i}$ respectively. 
Each vertex of $VG_{(\cdot)_i}$, $V_{(\cdot)_i}(c)$, represents the changes of the same vertex of different salience graphs. 
In other words, $V_{(\cdot)_i}(c)$ can measure the spatial variations of a specific category from a shallow layer to a deep layer. 
The edge of $VG_{(\cdot)_i}$ (i.e. $E_{(\cdot)_i}(c,k)$) is employed to evaluate the variations of inter-class correlation from a shallow layer to a deep layer. 

For transferring the knowledge of variations graphs, we devise variation graph loss function $\mathcal{L}_{vg}$ to mimic the discrepancy of variation graphs between $T$ and $S$. The objective function $\mathcal{L}_{vg}$ is defined as follows:
\begin{equation}
\begin{aligned}
     \mathcal{L}_{vg}=&\lambda_{1}\mathcal{L}_{vertex}+\lambda_{2} \mathcal{L}_{edge}\\=&\lambda_{1} \frac{1}{2\left| L\right|C}\sum_{i\in L}\sum_{c=1}^{C}\left\| V_{t_i}(c)-V_{s_i}(c)\right\|_{2}^{2}\\+&\lambda_{2}\frac{1}{2\left| L\right|C^{2}}\sum_{i\in L}\sum_{c=1}^{C}\sum_{k=1}^{C}\left\| E_{t_i}(c,k)-E_{s_i}(c,k)\right\|_{2}^{2}
     \label{equation7}
\end{aligned}    
\end{equation}
where $L$ notes the set of $(t_{i},t_{i^{'}})$ pair layers used to construct salience graphs.The $|L|$ is the size of L, which indicates the number of pair layers. We set $\left| L\right|=1$ in this paper.

\begin{figure}[t]
\begin{center}
\includegraphics[width=0.8\linewidth]{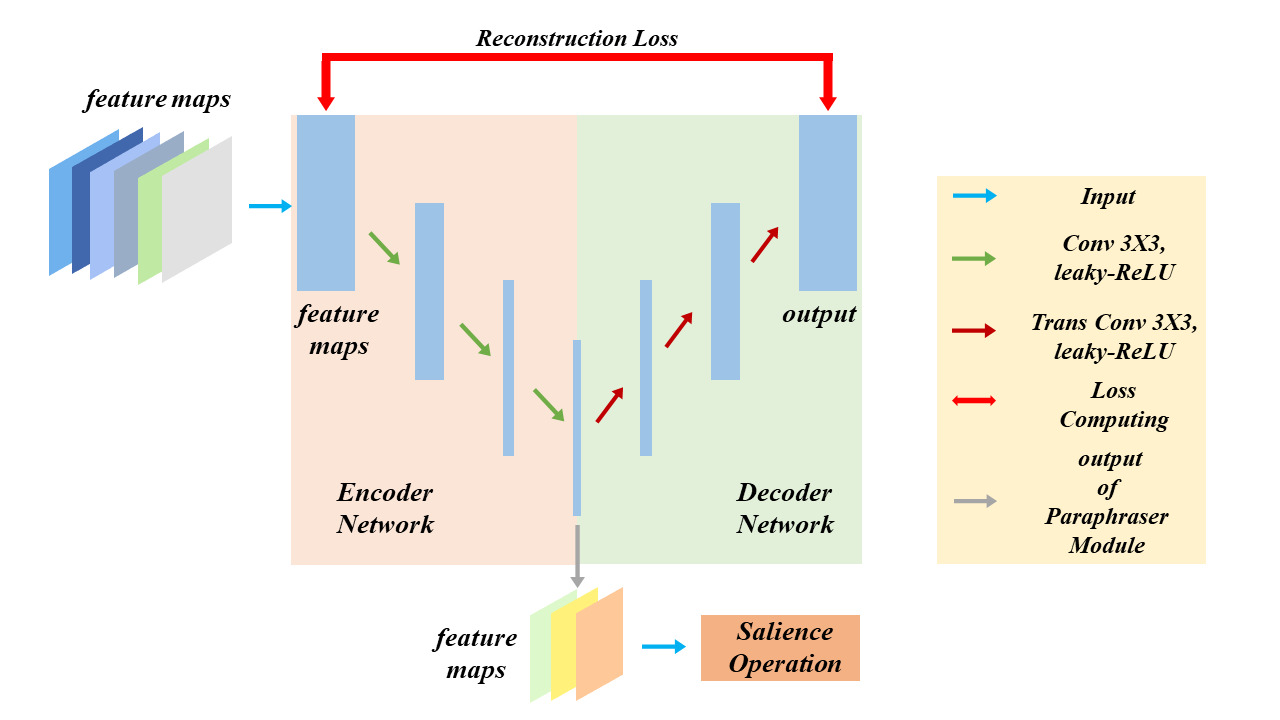}
\end{center}
   \vspace{-3mm}
   \caption{The framework of Paraphraser Module, which consists of two parts: a encoder network and a decoder network. The encoder network would be utilized to paraphrase knowledge of the teacher network.}
   \vspace{-3mm}
   \label{fig:paraphraser module}
\end{figure}
\subsection{Paraphraser Module}
It is worth noting that in our Graph Flow Distillation, the $F_{t_i}$ and $F_{s_i}$ always have different feature channels due to the different network architectures of $T$ and $S$. 
Thus, the knowledge embedded in $T$ can not be comprehended well by $S$, due to the different latent spaces of $F_{t_i}$ and $F_{s_i}$. 

Motivated by \cite{kim2018paraphrasing}, we utilize a Paraphraser Module $P_{m}$ to ease the latent space gap between $F_{t_i}$ and $F_{s_i}$ for refining the knowledge from $T$ and stabilizing the training of $S$. 
The overall architecture of our Paraphraser Module $P_{m}$ is illustrated in Fig.~\ref{fig:paraphraser module}. Specifically, our Paraphraser Module $P_{m}$ consists of a encoder network $P_{en}$ and a decoder network $P_{de}$. The encoder network $P_{en}$ is stacked by three $3\times 3$ convolutional operations with stride = 1 and padding = 1.  
The decoder network $P_{de}$ is stacked by three corresponding transposed convolutional layers. 
The leaky-ReLU is used after the convolutions and transposed convolutions. 
Given an input $F_{t_i}$, the encoder network $P_{en}$ aims to compress $F_{t_i}$ into the same latent space of $F_{s_i}$, while the decoder network $P_{de}$ is trained to reconstruct $F_{t_i}$ as much as possible. 
Therefore, the encoder network can maintain the informative knowledge. 
We utilize the reconstruction loss function $\mathcal{L}_{rec}$ to train our Paraphraser Module $P_{m}$, which is formulated as:
\begin{align}
    \mathcal{L}_{rec}=\left \|F_{t_i}-P_{m}(F_{t_i}) \right \|^{2}
\end{align}
Next, the trained encoder network $P_{en}$ can be exploited to embed $F_{t_i}$, which can be written as:
\begin{align}
    F_{t_{i}}^{'} = P_{en}(F_{t_i})
\end{align}
where $F_{t_{i}}^{'}$ is the output of $P_{en}$. 
Consequently, the construction of salience graphs of teacher networks could be modified as:
\begin{equation}
\begin{aligned}
     \Gamma_{t_i}&=MP_{en}(F_{t_i}) &=MF_{t_i}^{'}
\end{aligned}
\end{equation}
\subsection{Adversarial Distillation}
As depicted in Fig.~\ref{fig:framework}, we also adopt the adversarial training strategy \cite{xu2017training} to transfer the high-order knowledge hidden in the outputs of the teacher network $T$.
In our Adversarial Distillation, the discriminator network $D$ and the student network $S$ are alternatively updated in each training iteration. 

At first, the discriminator network $D$ is trained to distinguish the input from $S$ or $T$ for a given input which consists of the concatenation of $Y_{s}/Y_{t}$ and $X$. 
The training of the discriminator $D$ is constraint by adversarial loss function $\mathcal{L}_{adv}$, which is formulated as:
\begin{align}
    \mathcal{L}_{adv}=E_{Y_{s}\sim \mathcal{P}_{s}(Y_{s})}(D([Y_{s},X]))-E_{Y_{t}\sim \mathcal{P}_{t}(Y_{t})}(D([Y_{t},X])
\end{align}
where $[,]$ denotes the concatenation operation, and $E(\cdot)$ represents the expectation operation. The $\mathcal{L}_{adv}$ is designed to calculate the wasserstein distance between $T$ and $S$ \cite{liu2020structured}\cite{xu2017training}. Next, the student network $S$ approximates the output distribution of the teacher network $T$ as much as possible by maximizing $\mathcal{L}_{adv}$. 
%As the discriminator $D$ is fixed in the training of $S$, the $\mathcal{L}_{adv}$ for the optimization of $S$ can be modified as:
As the discriminator $D$ and $S$ are alternatively trained, and the D is fixed in the training of $S$, the $\mathcal{L}_{adv}$ for the optimization of $S$ can be modified as:
\begin{align}
    \mathcal{L}_{g}=E_{Y_{s}\sim \mathcal{P}_{s}(Y_{s})}(D([Y_{s},X])
\end{align}

\subsection{Logits Distillation}
Furthermore, we introduce Logits Distillation \cite{hinton2015distilling} to enforce the student network $S$ outputs more similar predictions with $T$. For the logits of $S$ or $T$, we produce a soft probability distribution $Z_{(\cdot)}$ by exploiting a softmax layer, which can be written as:
\begin{align}
    Z_{(\cdot)}(p)^m=\frac{exp(Y_{(\cdot)}(p)^m/\tau )}{\sum_{k=1}^{M}exp(Y_{(\cdot)}(p)^k/\tau )}
\end{align}
where $Z_{(\cdot)}(p)^m$ indicates the probability of $m_{th}$ class of $p_{th}$ pixel on $Y_{(\cdot)}$. The $\tau$ is a temperature which is set to 1. Since it can work well with the setting of 1, we do not extra adjust the hyperparameter. The $M$ is the total number of categories. The Kullback-Leibler divergence is employed to minimize the discrepancy between the teacher soft targets $Z_{t}$ and the student soft targets $Z_{s}$ as follows:
\begin{equation}
\begin{aligned}
    \mathcal{L}_{kd}&=\frac{1}{H\times W}\sum_{p=1}^{H\times W}KL(Z_s(p)||Z_t(p))\\&=\frac{1}{H\times W}\sum_{p=1}^{H\times W}\sum_{m=1}^{M} Z_{s}(p)^m log \frac{Z_{s}(p)^m }{Z_{t}(p)^m}
\end{aligned}    
\end{equation}
\subsection{Semi-supervised Learning}
Our method can improve the performance of compact networks without extra annotations compared with training from scratch.
It could provide a novel semi-supervised paradigm by distilling knowledge from cumbersome networks.
%Specifically, Our above three distillation methods can be implemented  
At first, we utilize annotated data to train our teacher network.
Next, the student network is trained by both labeled and unlabeled data.
Our student network is supervised under the groundtruth and our Graph Flow when the input images are annotated. 
On the contrary, our student network is only trained by our Graph Flow for unlabeled images.
%And for more detailed information about semi-supervised training, please refer to \emph{Section IV, Part F}
\subsection{Objective Functions}
In the optimization of our proposed Graph Flow, the training procedures can be implemented through three steps. 
At first, we employ $\mathcal{L}_{rec}$ to train $P_m$ while the parameters of pre-trained $T$ are frozen. Next, the discriminator $D$ is trained to discriminate $Y_{s}$ and $Y_{t}$ by minimizing $\mathcal{L}_{adv}$. At last, we optimize the student network $S$ with an overall objective function $\mathcal{L}$ that contains the basic cross-entropy loss $\mathcal{L}_{ce}$, variation graph loss $\mathcal{L}_{vg}$, adversarial loss $\mathcal{L}_{g}$, and logits distillation loss $\mathcal{L}_{kd}$. 
The $\mathcal{L}$ can be presented as:
\begin{equation}
    \begin{aligned}
        \mathcal{L}&=\mathcal{L}_{ce}+\mathcal{L}_{vg}-\lambda_{3}\mathcal{L}_{g}+\lambda_{4}\mathcal{L}_{kd}\\&=\mathcal{L}_{ce}+\lambda_{1}\mathcal{L}_{vertex}+\lambda_{2}\mathcal{L}_{edge}-\lambda_{3}\mathcal{L}_{g}+\lambda_{4}\mathcal{L}_{kd}
    \end{aligned}
\end{equation}
where the hyperparameters $\lambda_{1}$, $\lambda_{2}$, $\lambda_{3}$, $\lambda_{4}$ are set $10^{-5}$, $10^{-9}$, $0.1$, $1$ respectively. 

As for our semi-supervised strategy for dual efficient medical image segmentation, our $\mathcal{L}$ can be modified as
\begin{equation}
    \begin{aligned}
        \mathcal{L}&=\mathcal{L}_{ce_{labeled}}+\mathcal{L}_{vg}-\lambda_{3}\mathcal{L}_{g}+\lambda_{4}\mathcal{L}_{kd}\\&=\mathcal{L}_{ce_{labeled}}+\lambda_{1}\mathcal{L}_{vertex}+\lambda_{2}\mathcal{L}_{edge}-\lambda_{3}\mathcal{L}_{g}+\lambda_{4}\mathcal{L}_{kd}
    \end{aligned}
\end{equation}
where $labeled$ means that the cross-entropy loss is only used for labeled data. 
%And for more detailed information about semi-supervised training, please refer to \emph{Section IV, Part F}.
\section{Experiments}
\subsection{Datasets}
\textbf{\emph{1) Gastric Cancer Segmentation Dataset:}} The Gastric Cancer Segmentation Dataset\cite{sun2019accurate} (Gastric Cancer) consists of 500 pathological images (400 images for training and 100 images for validation) cropped from whole pathological slides of gastric regions with the resolution of $2048\times 2048$. The images in the dataset are manually annotated with two semantic categories: normal and cancerous area. Following our previous work\cite{zou2021coco}, we crop each $2048\times 2048$ image into four $1024\times 1024$ patches. 

\textbf{\emph{2) Synapse Multi-organ Segmentation Dataset:}} The Synapse Multi-organ Segmentation Dataset\cite{chen2021transunet} (Synapse) comes from MICCAI 2015 Multi-Atlas Abdomen Labeling Challenge, which contains 30 abdominal CT scans (18 CT scans for training and 12 CT scans for validation) with 3779 axial contrast-enhances abdominal CT images. The volume sizes of CT scans are from $85\times512\times512$ to $198\times512\times512$. 
We employ the preprocessed version of Synapse Multi-organ Segmentation Dataset offered by \cite{chen2021transunet} to conduct our experiments, which includes 8 abdominal organs: aorta, gallbladder, spleen, left kidney, right kidney, liver, pancreas, spleen, stomach. 

\textbf{\emph{3) Breast Ultrasound Images Dataset:}} The Breast Ultrasound Images Dataset \cite{al2020dataset} (BUSI) consists of 780 images with 437 benign cases, 210 benign masses, and 133 normal cases, which are collected from 600 female patients. Considering the main purpose of breast lesion segmentation in the clinical usage, we remove the normal cases that do not have breast lesion areas \cite{xue2021global}. 
We random split 518 images for training and 129 images for validation.
Since the images are different in their size, we resize the images with $256\times256$.

\textbf{\emph{4) CVC-ClinicDB Dataset:}} CVC-ClinicDB\cite{bernal2015wm} is a dataset containing 612 images which are extracted from 25 colonoscopy videos and selected 29 sequences from them.
Following \cite{fan2020pranet,kim2021uacanet}, we also add Kvasir \cite{jha2020kvasir} with 900 images into our training set.
Therefore, the total training set are extracted from CVC-ClinicDB and Kvasir containing 1450 images total.
The 62 images from CVC-ClinicDB are used for validation.
To unify the size of images, we resize the images with $256\times256$.

\subsection{Implementation Details}
We adopt the FANet\cite{li2020accurate} ($T1$) and TransUnet\cite{chen2021transunet} ($T2$) as our teacher networks respectively, in which the FANnet is an improved Unet based on self-attention mechanism and the TransUnet is a hybrid convolution and transformer architecture. 
Meanwhile, the lightweight Unet based on MobileNetv2\cite{sandler2018mobilenetv2} (Mobile U-Net) is employed as our student network $S$. Moreover, we also adopt the lightweight networks, ENet\cite{paszke2016enet} and ERFNet\cite{romera2017erfnet}, as our student networks.
The different teacher networks and student networks can examine the generalization of our method effectively.
We use Adam as the optimizer of $T1$, $S$, and $D$ respectively, with the weight decay = $0.0002$.
Following \cite{chen2021transunet}, the $T2$ are optimized by SGD optimizer with learning rate $0.01$, momentum $0.9$ and weight decay $0.0001$.
For training $P_{m}$, we employ the standard SGD as optimizer with the momentum = $0.9$, weight decay = $0.0002$. 
%The ``ploy" learning rate policy $(1-\frac{iter}{maxiter})^{power}$ is used in the optimizer, where the $power$ is set to $0.9$. 
The ``ploy" learning rate policy $(1-\frac{iter}{maxiter})^{power}$ is introduced in the training of each networks, where the $power$ is set to $0.9$. 
The initial learning rate is set to $0.003$ for optimizing $S, T1, P$ and 0.0002 for optimizing $D$, divided by 10 after every 50 epochs for $S, T1, P$.
In the training of overall networks, the epochs are set to 200 with the batch size of 8. 
Widely used data augmentation methods are also employed in our training \cite{zou2021coco}.
All experiments are completed on the Pytorch platform with $2$ NVIDIA TITAN XP GPUs. 
All experiments are repeated five times with different random seeds, and the results are presented in a format of means and standard deviations.
\begin{table}[ht]
\begin{center}
\caption{\label{tab:kd} Comparison with several representative knowledge distillation methods. CoCoD: CoCo DistillNet (our previous work). The digits on the upper left indicate the $p$ of the non-parametric Wilcoxon test between our method and others, and they are bolded if the $p\leqslant0.05$. The digits on the lower right indicate standard deviations.}
\resizebox{.5\textwidth}{!}{
\renewcommand{\arraystretch}{1.5} % default is 1.0
\begin{tabular}{cc|cc|cc}
\hline
\multicolumn{2}{c|}{\multirow{2}{*}{\begin{tabular}[c]{@{}c@{}}Knowledge \\  Distillation\end{tabular}}} & \multicolumn{2}{c|}{Gastric Cancer} & \multicolumn{2}{c}{Synapse}        \\ \cline{3-6} 
\multicolumn{2}{c|}{}                                                                                       & ACC$\uparrow$              & mIOU$\uparrow$             & average DSC$\uparrow$     & average HD$\downarrow$      \\ \hline
\multicolumn{1}{c|}{T1: FANet}                               & \multirow{2}{*}{Years}                        & 0.9005$_{\pm0.001}$           & 0.8189$_{\pm0.002}$           & 0.7950$_{\pm0.003}$          & 24.5175$_{\pm0.510}$           \\
\multicolumn{1}{c|}{S: Mobile U-Net}                        &                                               & 0.8544$_{\pm0.002}$           & 0.7456$_{\pm0.001}$           & 0.7331$_{\pm0.003}$          & 40.7060$_{\pm3.886}$           \\ \hline
\multicolumn{1}{c|}{KD\cite{hinton2015distilling}}                                     & 2015                                          & $^{\textbf{0.008}}\textrm{0.8608}_{\pm0.001}$          & $^{\textbf{0.004}}\textrm{0.7554}_{\pm0.002}$           & $^{\textbf{0.014}}\textrm{0.7719}_{\pm0.004}$          & $^{\textbf{0.031}}\textrm{34.8758}_{\pm6.187}$        \\
\multicolumn{1}{c|}{AT\cite{zagoruyko2016paying}}                                     & 2016                                          & $^{\textbf{0.003}}\textrm{0.8603}_{\pm0.003}$           & $^{\textbf{0.007}}\textrm{0.7548}_{\pm0.004}$           & $^{\textbf{0.001}}\textrm{0.7449}_{\pm0.005}$          & $^{\textbf{0.016}}\textrm{37.9143}_{\pm3.605}$          \\
\multicolumn{1}{c|}{FSP\cite{yim2017gift}}                                    & 2017                                          & $^{\textbf{0.015}}\textrm{0.8664}_{\pm0.001}$            & $^{\textbf{0.029}}\textrm{0.7643}_{\pm0.002}$           & $^{\textbf{0.004}}\textrm{0.7685}_{\pm0.007}$          & $^{\textbf{0.008}}\textrm{33.8504}_{\pm5.402}$          \\
\multicolumn{1}{c|}{FT\cite{kim2018paraphrasing}}                                     & 2018                                          & $^{\textbf{0.013}}\textrm{0.8668}_{\pm0.001}$           & $^{\textbf{0.005}}\textrm{0.7650}_{\pm0.001}$           & $^{\textbf{0.004}}\textrm{0.7619}_{\pm0.001}$          & $^{\textbf{0.011}}\textrm{33.2281}_{\pm3.434}$          \\
\multicolumn{1}{c|}{VID\cite{ahn2019variational}}                                    & 2019                                          & $^{\textbf{0.004}}\textrm{0.8601}_{\pm0.002}$           & $^{\textbf{0.008}}\textrm{0.7546}_{\pm0.003}$           & $^{\textbf{0.027}}\textrm{0.7790}_{\pm0.001}$          & $^{\textbf{0.039}}\textrm{31.7290}_{\pm3.065}$         \\
\multicolumn{1}{c|}{SKD\cite{liu2020structured}}                                    & 2020                                          & $^{\textbf{0.044}}\textrm{0.8740}_{\pm0.002}$           & $^{0.110}\textrm{0.7761}_{\pm0.003}$           & $^{\textbf{0.016}}\textrm{0.7803}_{\pm0.003}$         & $^{\textbf{0.004}}\textrm{33.8104}_{\pm8.361}$          \\
\multicolumn{1}{c|}{IFVD\cite{wang2020intra}}                                   & 2020                                          & $^{\textbf{0.015}}\textrm{0.8670}_{\pm0.001}$           & $^{\textbf{0.014}}\textrm{0.7651}_{\pm0.002}$           & $^{\textbf{0.023}}\textrm{0.7694}_{\pm0.004}$          & $^{\textbf{0.016}}\textrm{37.0198}_{\pm5.555}$         \\ 
\multicolumn{1}{c|}{EMKD\cite{qin2021efficient}}                                   & 2021                                          & $^{\textbf{0.027}}\textrm{0.8733}_{\pm0.002}$           & $^{\textbf{0.041}}\textrm{0.7751}_{\pm0.003}$           & $^{\textbf{0.003}}\textrm{0.7713}_{\pm0.006}$          & $^{0.151}\textrm{30.8120}_{\pm2.466}$         \\ \hline
\multicolumn{1}{c|}{CoCoD\cite{zou2021coco}}                                  & 2021                                          & $^{0.050}\textrm{0.8779}_{\pm0.001}$           & $^{\textbf{0.029}}\textrm{0.7823}_{\pm0.001}$           & $^{0.212}\textrm{0.7811}_{\pm0.004}$          & $^{\textbf{0.002}}\textrm{35.1829}_{\pm7.295}$          \\
\multicolumn{1}{c|}{Graph Flow}                             &                                               & $\textbf{0.8869}_{\pm0.001}$  & \textbf{0.7967$_{\pm0.001}$}  & \textbf{0.7870$_{\pm0.002}$} & \textbf{29.0594$_{\pm0.599}$} \\ \hline
\end{tabular}
}
\end{center}
\end{table}
\begin{figure}[h]
\begin{center}
\includegraphics[width=1.0\linewidth]{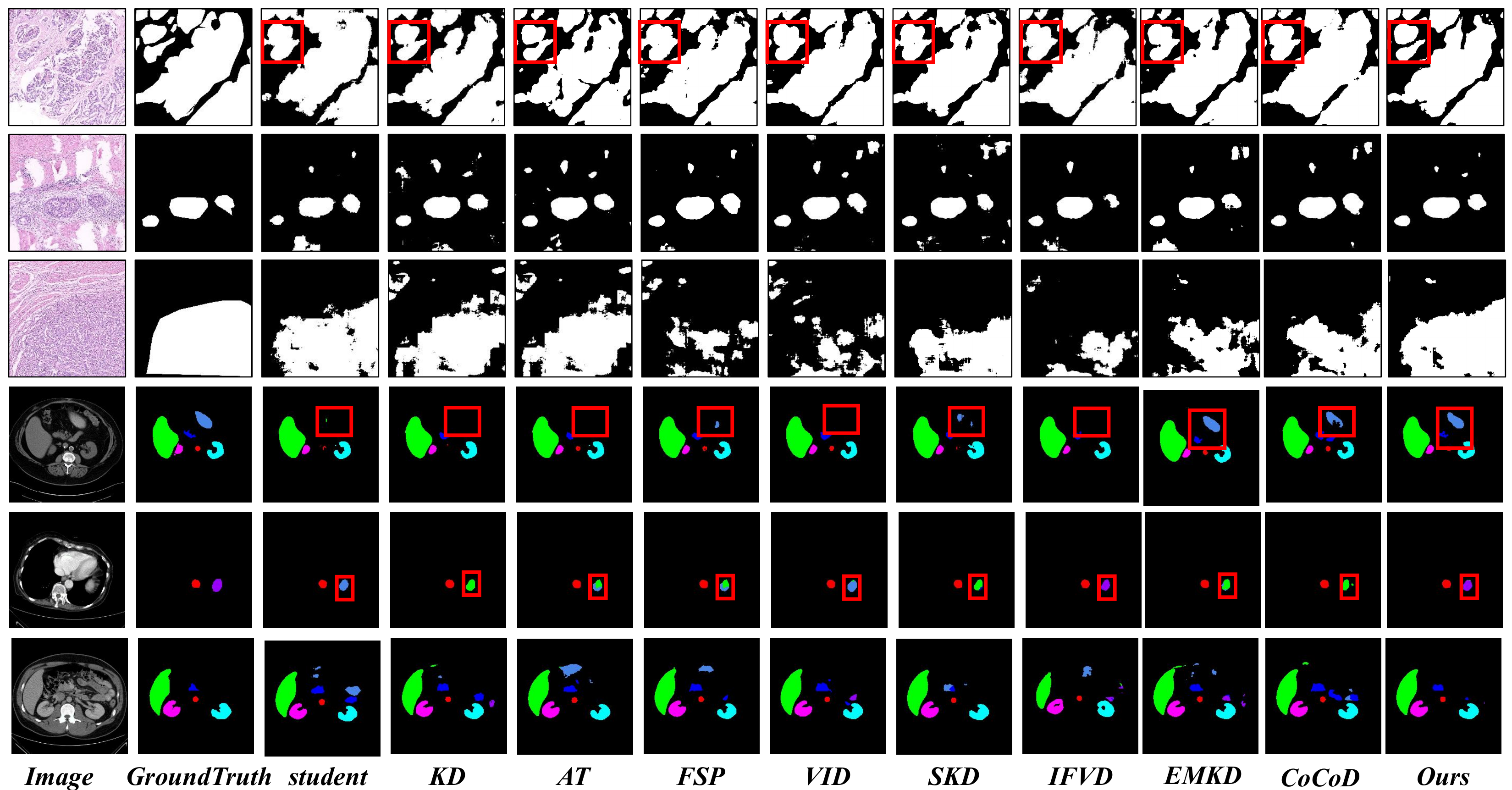}
\end{center}
   \vspace{-3mm}
   \caption{Visualized comparisons with several knowledge distillation methods on Gastric Cancer and Synapse. \textbf{The same color rectangles signify noteworthy distinctions among these methods, and the other rectangles in the rest of figures in this paper are the same.} \textbf{Zoom in} for better details.}
   \vspace{-3mm}
   \label{fig:kd_visualization}
\end{figure}
\begin{table*}[t]
\begin{center}
\caption{\label{tab:patchsize} Ablation results of Graph Flow Distillation based on different patch-wise salience regions on Gastric Cancer and Synapse. We also report DSC of every abdominal organ except average DSC and HD on Synapse. $64\times64$ refers to the size of total feature maps. }
\resizebox{\textwidth}{!}{
\renewcommand{\arraystretch}{1.5} % default is 1.0
\begin{tabular}{ccccccccccccc}
\hline
\multicolumn{1}{c|}{\multirow{2}{*}{Methods}} & \multicolumn{2}{c|}{Gastric Cancer}           & \multicolumn{10}{c}{Synapse}                                                                                                                                                                   \\ \cline{2-13} 
\multicolumn{1}{c|}{}                  & ACC $\uparrow$             & \multicolumn{1}{c|}{mIOU $\uparrow$}            & average DSC $\uparrow$             & \multicolumn{1}{c|}{average HD $\downarrow$}               & aorta $\uparrow$           & gallbladder $\uparrow$     & left kidney $\uparrow$     & right kidney $\uparrow$    & liver $\uparrow$           & pancreas $\uparrow$        & spleen $\uparrow$          & stomach $\uparrow$         \\ \hline
\multicolumn{1}{c|}{T1: FANet}          & 0.9005$_{\pm0.001}$          & \multicolumn{1}{c|}{0.8189$_{\pm0.002}$}          & 0.7950$_{\pm0.003}$           & \multicolumn{1}{c|}{24.5175$_{\pm0.510}$}          & 0.8725$_{\pm0.007}$          & 0.6273$_{\pm0.008}$          & 0.8145$_{\pm0.014}$          & 0.8153$_{\pm0.010}$          & 0.9523$_{\pm0.001}$          & 0.6092$_{\pm0.003}$          & 0.8985$_{\pm0.001}$         & 0.7630$_{\pm0.010}$           \\ \hline
\multicolumn{1}{c|}{S: Mobile U-Net}   & 0.8544$_{\pm0.002}$          & \multicolumn{1}{c|}{0.7456$_{\pm0.001}$ }          & 0.7331$_{\pm0.003}$          & \multicolumn{1}{c|}{40.7060$_{\pm3.886}$}          & 0.8367$_{\pm0.003}$          & 0.4247$_{\pm0.051}$          & 0.7986$_{\pm0.004}$          & 0.7896$_{\pm0.008}$          & 0.9439$_{\pm0.004}$          & 0.4951$_{\pm0.017}$          & 0.8703$_{\pm0.006}$           & 0.7059$_{\pm0.004}$          \\ \hline
\multicolumn{13}{c}{\textbf{Graph Flow Distillation (Graph) with different sizes of Patch-wise Salience Regions}}                                                                      \\ \hline
\multicolumn{1}{c|}{+Graph 64$\times$64}  & 0.8745$_{\pm0.001}$          & \multicolumn{1}{c|}{0.7868$_{\pm0.002}$}         & 0.7708$_{\pm0.002}$          & \multicolumn{1}{c|}{30.7699$_{\pm1.063}$}          & 0.8514$_{\pm0.006}$         & 0.5873$_{\pm0.041}$         & 0.8263$_{\pm0.025}$          & 0.7972$_{\pm0.021}$          & 0.9443$_{\pm0.002}$          & 0.5435$_{\pm0.021}$ & 0.8883$_{\pm0.011}$          & 0.7278$_{\pm0.012}$          \\ \hline
\multicolumn{1}{c|}{+Graph 3$\times$3}                 & 0.8737$_{\pm0.001}$          & \multicolumn{1}{c|}{0.7757$_{\pm0.004}$}          & \textbf{0.7829$_{\pm0.001}$} & \multicolumn{1}{c|}{\textbf{26.9907$_{\pm4.138}$}} & 0.8511$_{\pm0.021}$          & \textbf{0.6274$_{\pm0.013}$} & \textbf{0.8462$_{\pm0.012}$} & \textbf{0.8140$_{\pm0.011}$}          & 0.9451$_{\pm0.005}$ & 0.5449$_{\pm0.013}$         & \textbf{0.8982$_{\pm0.017}$}          & 0.7375$_{\pm0.017}$         \\ \hline
\multicolumn{1}{c|}{+Graph 5$\times$5}                 & 0.8771$_{\pm0.005}$          & \multicolumn{1}{c|}{0.7810$_{\pm0.001}$}          & 0.7751$_{\pm0.003}$          & \multicolumn{1}{c|}{28.1392$_{\pm1.380}$}          & \textbf{0.8540$_{\pm0.005}$}          & 0.5906$_{\pm0.019}$          & 0.8045$_{\pm0.022}$          & 0.7842$_{\pm0.021}$          & 0.9440$_{\pm0.002}$          & 0.5907$_{\pm0.020}$          & 0.8940$_{\pm0.003}$         & 0.7391$_{\pm0.016}$          \\ \hline
\multicolumn{1}{c|}{+Graph 7$\times$7}  & 0.8745$_{\pm0.001}$          & \multicolumn{1}{c|}{0.7770$_{\pm0.001}$}           & 0.7735$_{\pm0.002}$           & \multicolumn{1}{c|}{25.8437$_{\pm2.586}$ }          & 0.8343$_{\pm0.007}$          & 0.6066$_{\pm0.022}$          & 0.8410$_{\pm0.006}$          & 0.8078$_{\pm0.024}$          & \textbf{0.9460$_{\pm0.002}$}          & 0.5269$_{\pm0.018}$          & 0.8866$_{\pm0.006}$          & 0.7386$_{\pm0.028}$ \\ \hline
\multicolumn{1}{c|}{+Graph 9$\times$9}                 & \textbf{0.8807$_{\pm0.002}$} & \multicolumn{1}{c|}{\textbf{0.7868$_{\pm0.003}$}} & 0.7752$_{\pm0.002}$         & \multicolumn{1}{c|}{31.5875$_{\pm2.533}$}          & 0.8501$_{\pm0.024}$ & 0.6108$_{\pm0.040}$          & 0.8186$_{\pm0.010}$          & 0.7856$_{\pm0.011}$          & 0.9447$_{\pm0.002}$          & \textbf{0.5604$_{\pm0.021}$}          & 0.8882$_{\pm0.008}$        & \textbf{0.7393$_{\pm0.008}$}          \\ \hline
\end{tabular}}
\vspace{-5mm}
\end{center}
\end{table*}
\vspace{-3mm}
\subsection{Evaluation Metrics}
We employ different evaluation metrics to verify the performance of our method on Gastric Cancer and Synapse datasets due to their different modalities. 
\textbf{(1) Gastric Cancer Segmentation Dataset:} Following \cite{sun2019accurate} and \cite{zou2021coco}, we report the pixel accuracy (ACC) and the mean Intersection-over-Union (mIoU) of our results for evaluation.
\textbf{(2) Synapse Multi-organ Segmentation Dataset:} Following \cite{chen2021transunet}, we employ the average Dice Similarity Coefficient (DSC) and the average Hausdorff Distance (HD) of the 8 abdominal organs. 
\textbf{(3) Breast Ultrasound Images and CVC-ClinicDB Datasets:}
Folllowing \cite{xue2021global, kim2021uacanet}, we employ the average Dice Similarity Coefficient (DSC) and the mean Intersection-over-Union (mIoU) of our results for evaluation.

Furthermore, We also compute the sum of point operations (FLOPs) and the number of network parameters (Params) to measure the model complexity.
Besides, the non-parametric Wilcoxon test \cite{valvano2021learning} are conducted to evaluate if improvements of our method are statistically signifificance with $p\leqslant0.05$.

\subsection{Results of Comparative Experiments}
\textbf{\emph{1) Results of the Comparisons with Other Knowledge Distillation Methods:}} We demonstrate the superiority of our method by comparing with the state-of-the-art popular knowledge distillation methods, as shown in Table.~\ref{tab:kd}. 
Moreover, the visualization in Fig.~\ref{fig:kd_visualization} also shows the significant advantages of our method.
%From the Table.~\ref{tab:kd}, we can conclude that our method has statistically significant improvement compared with most state-of-the-art popular knowledge distillation.
Since KD\cite{hinton2015distilling}, AT\cite{zagoruyko2016paying}, FT\cite{kim2018paraphrasing} and VID\cite{ahn2019variational} all focus on the mimicking of feature maps or the transformation of themselves from middle layer or outputs, they exhibit similar performance and our method surpass them almost $3\%$ in mIOU on Gastric Cancer.
%Compared with FSP\cite{yim2017gift}, our method can better eliminate redundant information contained in the flow of different layers, which results to $3.24\%$ increase in mIOU on Gastric Cancer.} 
Compared with FSP\cite{yim2017gift}, our method suppresses redundant activations except for patch-wise salience regions, which are unrelated to semantic-aware knowledge in deep layers, and focuses more on structural semantic-aware knowledge (class-wise spatial information and inter-class correlation) of an intermediate layer, which results to 2.05\% increase in mIOU on Gastric Cancer.
In addition, our method has significant reduction of HD on Synapse comparing with SKD\cite{liu2020structured} and IFVD\cite{wang2020intra}.
%The lower Hausdorff Distance means the shape of our predictions closer with the annotations.
%It indicates that our method can better represent holistic semantic-aware knowledge, which is crucial for medical image segmentation.  
Moreover, our method can improve $1.36\%$ ACC on Gastric Cancer and $1.57\%$ average DSC on Synapse compared with EMKD, which demonstrate that our method can perform better than EMKD \cite{qin2021efficient} for medical image segmentation.
%From the perspective of generalization and robustness of the algorithms, we find that the above-mentioned KD methods have limitations at the cross-modality implementations as reported in Table I. 
%On the contrary, our Graph Flow exhibits prominent performance on the both datasets with cross-modality images obtained from pathological slides and CT scanners, respectively.
%It indicates that our method has better generalization ability for different modalities of images. 
%Overall, our method achieves the state-of-the-art performance. 

\begin{figure}[ht]
\begin{center}
\includegraphics[width=1\linewidth]{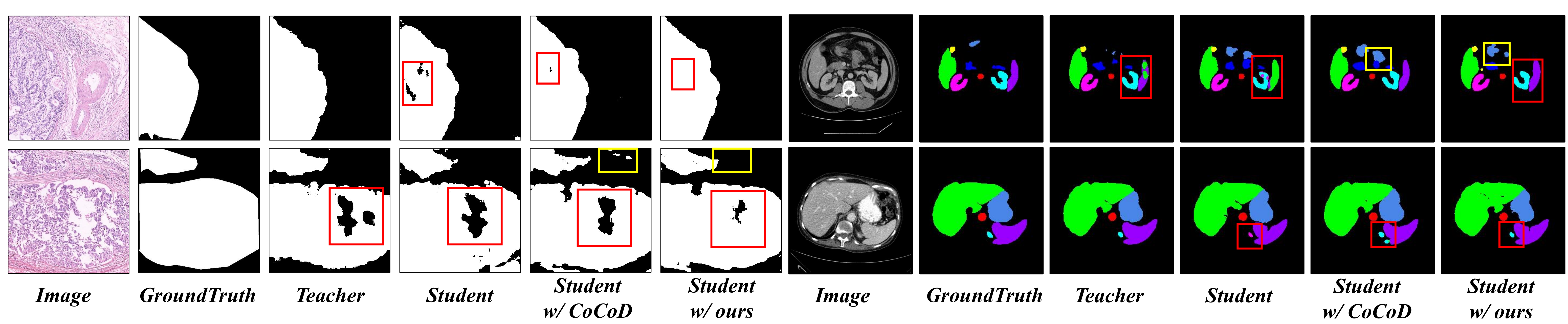}
\end{center}
   \vspace{-3mm}
   \caption{Visualized comparisons with our prior work on Gastric Cancer and Synapse. CoCoD: CoCo DistillNet (our previous work). Left: Gastric Cancer. Right: Synapse. \textbf{Zoom in} for better details.}
   \vspace{-6mm}
   \label{fig:visualization}
\end{figure}

\textbf{\emph{2) Results of the Comparisons with Our Prior Work:}} To demonstrate the comprehensive extension of our \textbf{Graph Flow} with our preliminary CoCo DistillNet\cite{zou2021coco}, we conduct quantitative and qualitative analysis, as shown in the last two rows of Table.~\ref{tab:kd} and Fig.~\ref{fig:visualization}.
Quantitatively, our method yields remarkable improvement compared to our previous work, especially in HD on Synapse with the $6.1235$ mm reduction.
On Gastric Cancer, our method achieve the progresses of $0.9\%$ on ACC and $1.44\%$ on mIOU.
%It demonstrates our predictions are shape-closer with the annotations, as shown in the last row in Fig.~\ref{fig:visualization}.
%For the progresses of $0.9\%$ on ACC and $1.44\%$ on mIOU, 
%it means that our method could better eliminate extra error predictions better than our previous work as shown in Fig.~\ref{fig:visualization} left.
%It indicates that our method defines a better refined knowledge to represent the flow between different layers than our previous work. }
Qualitatively, we can observe that the student network trained by \textbf{Graph Flow} is not influenced by the surrounding cancerous areas. 
Our student network could predict more correct normal areas (yellow boxes in second row left), and could distinguish different organs (red boxes in first row right).
Moreover, the student trained by Graph Flow can discriminate left and right kidney (red boxes in the last row right) even without the discrepancy of spatial position (only a kidney exists in this slice).
Furthermore, our method can help the student network avoid some error-prone details, as shown in the red boxes in last row left of Fig.~\ref{fig:visualization}.
%We conjecture that our method can teach student to capture the spatial relation between different categories because our salience graph can encode class-wise spatial information. 
%Moreover, our method also transfers the flow of inter-class correlation from teacher to student. Therefore, the student trained by Graph Flow can discriminate left and right kidney (red boxes in the last row right) even without the discrepancy of spatial position (only a kidney exists in this slice).
%Furthermore, our Graph Flow can better cooperate with the supervision of groundtruth than our prior CoCo DistillNet, which can help the student network avoid the mistakes of the teacher, as shown in the red boxes in last row left of Fig.~\ref{fig:visualization}.
%\textcolor{blue}{On the other hand, we avoid to compute the Channel Mixed Spatial Similarity Module (CMSS) proposed by our previous work, which involves the resolution-wise multiplication operations ($B \times HW\times HW$) with the huge storage memory.
%Therefore, our Graph Flow does not add complexity and needs less storage memory than our previous work with a better performance.} 

%\vspace{-3mm}

\vspace{-3mm}
\subsection{Results of Ablation Study}
\textbf{\emph{1) Results of the Patch-wise Salience Region Ablation:}} To report the effect of different sizes on salience regions $P$, we conduct patch-wise salience region ablation on both Gastric Cancer and Synapse. 
%As shown in Table.~\ref{tab:patchsize}, the comparisons between local and non-local ($64\times64$) salience regions demonstrate that our salience operation can eliminate redundant knowledge, thus the salience graph could focus more on favorable semantic class-aware knowledge.
%\textcolor{blue}{Furthermore, the patch size $9\times9$ can gain the best performance compared with other patch sizes on Gastric Cancer, which can improve the performance of $S$ without Graph Flow Distillation by $2.63\%$ (ACC) and $4.12\%$ (mIOU) respectively.}
As shown in Table.~\ref{tab:patchsize}, the patch size $9\times9$ can gain the best performance compared with other patch sizes on Gastric Cancer, which can improve the performance of $S$ without Graph Flow Distillation by $2.63\%$ (ACC) and $4.12\%$ (mIOU) respectively.
This can be explained by the large circular characteristics of the gastric cancer areas in pathological image. 
For Synapse, we observe that the patch size $3\times3$ outperforms $S$ without distillation by $4.98\%$ (average DSC) and $13.7153$ mm (average HD) respectively.
This attributes to that Synapse has many small abdominal organs (e.g. aorta and gallbladder), and too large size of $P$ causes the salience graph to obtain redundant and irrelevant class-wise information in certain channels when encoding small abdominal organs.
In other words, the small abdominal organs are sensitive to the size of $P$, but the large abdominal organs are relative insensitive. 
For example, as shown in Table.~\ref{tab:patchsize}, the range of average DSCs of gallbladder can achieve $4.01\%$ on different patch sizes. 
On the contrary, the range of average DSCs of liver only has $0.20\%$.
%Therefore, the dataset tends to select small patch-wise salience regions of $P$, if there are great differences in category size (e.g. Synapse). 
%On the other hand, the dataset with resemblance in category size (e.g. Gastric Cancer) may employ relative large $P$. 
%In our all experiments, we select $9\times9$ for Gastric Cancer and $3\times3$ for Synapse.
In our all experiments, we select $9\times9$ for Gastric Cancer, BUSI, and CVC-ClinicDB Dataset, and $3\times3$ for Synapse.
\begin{table*}[ht]
\begin{center}
\caption{\label{tab:student} The results of our Graph Flow on different teacher and student networks.  w/o and w/ represents the training of $S$ with Graph Flow or without Graph Flow respectively.}
\resizebox{\textwidth}{!}{
\renewcommand{\arraystretch}{1.5} % default is 1.0
\begin{tabular}{cl|cc|cc|cc|cc|c|c}
\hline
\multicolumn{2}{c|}{\multirow{2}{*}{Networks}}                                           & \multicolumn{2}{c|}{Gastric Cancer} & \multicolumn{2}{c|}{Synapse} & \multicolumn{2}{c|}{BUSI} &  \multicolumn{2}{c|}{CVC-ClinicDB} & \multicolumn{1}{l|}{\multirow{2}{*}{FLOPs(G)}} & \multicolumn{1}{l}{\multirow{2}{*}{Params(M)}} \\ \cline{3-10}
\multicolumn{2}{c|}{}                                                                    & ACC$\uparrow$              & mIOU$\uparrow$             & average DSC$\uparrow$   & average HD$\downarrow$  &  average DSC$\uparrow$ &  mIOU$\uparrow$ &  average DSC$\uparrow$ &  mIOU$\uparrow$  &\multicolumn{1}{l|}{}                          & \multicolumn{1}{l}{}                           \\ \hline
\multicolumn{2}{c|}{T1: FANet}                                                            &  0.9005$_{\pm0.001}$            & 0.8189$_{\pm0.002}$            & 0.7950$_{\pm0.003}$        & 24.5175$_{\pm0.510}$    & 0.8038$_{\pm0.004}$  & 0.8173$_{\pm0.003}$  & 0.9006$_{\pm0.016}$  & 0.9015$_{\pm0.014}$ & 171.556                                        & 38.250                                         \\ \hline
\multicolumn{2}{c|}{T2: TransUnet}                                                            &  0.8830$_{\pm0.006}$            & 0.7905$_{\pm0.009}$            & 0.7786$_{\pm0.003}$        & 28.2066$_{\pm0.356}$    & 0.7949$_{\pm0.002}$  & 0.8085$_{\pm0.001}$  & 0.9257$_{\pm0.010}$  & 0.9245$_{\pm0.009}$ & 24.610                                        & 105.281                                        \\ \hline
\multicolumn{1}{c|}{\multirow{3}{*}{Mobile U-Net}} & \multicolumn{1}{c|}{w/o}                       & 0.8544$_{\pm0.002}$             & 0.7456$_{\pm0.001}$            & 0.7331$_{\pm0.003}$        & 40.7060$_{\pm3.886}$       & 0.7748$_{\pm0.010}$ & 0.7937$_{\pm0.006}$ &0.8555$_{\pm0.003}$ &0.8622$_{\pm0.003}$ & \multirow{3}{*}{1.492}                         & \multirow{3}{*}{4.640}                         \\
\multicolumn{1}{c|}{}                              & T1: w/                        & \textbf{0.8869$_{\pm0.001}$}           & \textbf{0.7967$_{\pm0.001}$}            & \textbf{0.7870$_{\pm0.002}$}      & \textbf{29.0594$_{\pm0.599}$}     & \textbf{0.7971$_{\pm0.011}$} & \textbf{0.8124$_{\pm0.009}$} & \textbf{0.8792$_{\pm0.003}$} & \textbf{0.8817$_{\pm0.002}$} &                                                &                                                \\ 
\multicolumn{1}{c|}{}                              & T2: w/                        & \textbf{0.8629$_{\pm0.001}$}           & \textbf{0.7588$_{\pm0.001}$}            & \textbf{0.7748$_{\pm0.006}$}      & \textbf{31.2980$_{\pm1.701}$}     & \textbf{0.7908$_{\pm0.008}$} & \textbf{0.8076$_{\pm0.006}$} & \textbf{0.8675$_{\pm0.005}$} & \textbf{0.8721$_{\pm0.005}$} &                                                &                                                \\ \hline
\multicolumn{1}{c|}{\multirow{3}{*}{ENet}}         & \multicolumn{1}{c|}{w/o}                       & 0.8680$_{\pm0.001}$           & 0.7667$_{\pm0.002}$           & 0.7440$_{\pm0.008}$       & 31.7275$_{\pm3.592}$  & 0.7689$_{\pm0.019}$  & 0.7917$_{\pm0.014}$  & 0.8614$_{\pm0.006}$ & 0.8668$_{\pm0.005}$ & \multirow{3}{*}{0.516}                         & \multirow{3}{*}{0.349}                         \\
\multicolumn{1}{c|}{}                              & T1: w/                        &\textbf{0.8847$_{\pm0.001}$}           & \textbf{0.7932$_{\pm0.002}$}           & \textbf{0.7637$_{\pm0.002}$}        & \textbf{27.0532$_{\pm2.766}$}   & \textbf{0.7992$_{\pm0.006}$} &\textbf{0.8140$_{\pm0.005}$} &\textbf{0.8786$_{\pm0.004}$} &\textbf{0.8810$_{\pm0.003}$} &                                                &                                                \\ 
\multicolumn{1}{c|}{}                              & T2: w/                        &\textbf{0.8799$_{\pm0.002}$}           & \textbf{0.7855$_{\pm0.003}$}           & \textbf{0.7602$_{\pm0.002}$}        & \textbf{30.8132$_{\pm1.508}$}   & \textbf{0.7907$_{\pm0.005}$} &\textbf{0.8077$_{\pm0.003}$} &\textbf{0.9006$_{\pm0.008}$} &\textbf{0.9009$_{\pm0.007}$} &                                                &                                                \\ \hline
\multicolumn{1}{c|}{\multirow{3}{*}{ERFNet}}       & \multicolumn{1}{c|}{w/o } & 0.8655$_{\pm0.003}$          & 0.7629$_{\pm0.005}$           & 0.7463$_{\pm0.007}$   & 34.0169$_{\pm3.550}$  & 0.7732$_{\pm0.018}$ & 0.7938$_{\pm0.013}$ & 0.8625$_{\pm0.003}$  &0.8679$_{\pm0.003}$ & \multirow{3}{*}{3.681}                         & \multirow{3}{*}{2.063}                         \\
\multicolumn{1}{c|}{}                              & T1: w/                        & \textbf{0.8878$_{\pm0.001}$ }           & \textbf{0.7982$_{\pm0.002}$}           & \textbf{0.7715$_{\pm0.003}$}        & \textbf{30.0776$_{\pm3.300}$}   & \textbf{0.7994$_{\pm0.004}$} &\textbf{0.8125$_{\pm0.003}$} &\textbf{0.8890$_{\pm0.010}$} &\textbf{0.8908$_{\pm0.010}$} &                                                &                                                \\ 
\multicolumn{1}{c|}{}                              & T2: w/                        & \textbf{0.8781$_{\pm0.002}$ }           & \textbf{0.7827$_{\pm0.004}$}           & \textbf{0.7626$_{\pm0.002}$}        & \textbf{29.6510$_{\pm0.957}$}   & \textbf{0.7890$_{\pm0.007}$} &\textbf{0.8049$_{\pm0.006}$} &\textbf{0.9001$_{\pm0.004}$} &\textbf{0.9007$_{\pm0.004}$} &                                                &                                                \\\hline
\end{tabular}}
\end{center}
\vspace{-3mm}
\end{table*}

\begin{table}[ht]
\vspace{-3mm}
\begin{center}
\caption{\label{tab:components} Ablation study of the components of our Graph Flow. Graph: Graph Flow Distillation. Adv: Adversarial Distillation. Logits: Logits Distillation.}
\resizebox{.5\textwidth}{!}{
\renewcommand{\arraystretch}{1.5} % default is 1.0
\begin{tabular}{cll|cc|cc}
\hline
\multicolumn{3}{c|}{\multirow{2}{*}{Methods}} & \multicolumn{2}{c|}{Gastric Cancer} & \multicolumn{2}{c}{Synapse} \\ \cline{4-7} 
\multicolumn{3}{c|}{}                                 & ACC $\uparrow$                  & mIOU $\uparrow$                & average DSC $\uparrow$       & average HD $\downarrow$       \\ \hline
\multicolumn{3}{c|}{T1: FANet}                         & 0.9005$_{\pm0.001}$                & 0.8189$_{\pm0.002}$             & 0.7950$_{\pm0.003}$            & 24.5175$_{\pm0.510}$          \\ \hline
\multicolumn{3}{c|}{S: Mobile U-Net}                  & 0.8544$_{\pm0.002}$                & 0.7456$_{\pm0.001}$               & 0.7331$_{\pm0.003}$            & 40.7060$_{\pm3.886}$           \\
\multicolumn{3}{c|}{+Graph}                           & 0.8807$_{\pm0.002}$                & 0.7868$_{\pm0.003}$             & 0.7829$_{\pm0.001}$            & \textbf{26.9907$_{\pm4.138}$} \\
\multicolumn{3}{c|}{+Graph+Adv}                       & 0.8851$_{\pm0.001}$                & 0.7938$_{\pm0.002}$               & 0.7854$_{\pm0.003}$            & 27.9895$_{\pm4.056}$          \\
\multicolumn{3}{c|}{+Graph+Adv+Logits}                & \textbf{0.8869$_{\pm0.001}$}      & \textbf{0.7967$_{\pm0.001}$}     & \textbf{0.7870$_{\pm0.002}$}  & 29.0594$_{\pm0.599}$         \\ \hline
\end{tabular}}
\vspace{-4mm}
\end{center}
\end{table}
\begin{figure}[ht]
\begin{center}
\vspace{1mm}
\includegraphics[width=0.8\linewidth]{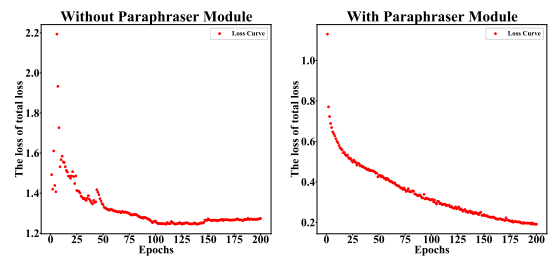}
\end{center}
   \vspace{-3mm}
   \caption{The loss curves of ablation analysis on Graph Flow without or with Paraphraser Module.}
   \vspace{-4mm}
   \label{fig:loss_compare}
\end{figure}
\begin{table}[ht]
\begin{center}
\vspace{-3mm}
\caption{\label{tab:Paraphraser} The examinations of effectiveness with our Paraphraser Module. CoCoD: CoCo DistillNet. PM:Paraphraser Module.}
\resizebox{.5\textwidth}{!}{
\renewcommand{\arraystretch}{1.5} % default is 1.0
\begin{tabular}{c|cc|cc}
\hline
\multirow{2}{*}{Methods} & \multicolumn{2}{c|}{Gastric Cancer} & \multicolumn{2}{c}{Synapse} \\ \cline{2-5} 
                        & ACC$\uparrow$              & mIOU$\uparrow$             & avergae DSC$\uparrow$  & avergage HD$\downarrow$  \\ \hline
T1: FANet                & 0.9005$_{\pm0.001}$           & 0.8189$_{\pm0.002}$            & 0.7950$_{\pm0.003}$       & 24.5175$_{\pm0.510}$      \\ \hline
S:Mobile U-Net          & 0.8544$_{\pm0.002}$            & 0.7456$_{\pm0.001}$           & 0.7331$_{\pm0.003}$       & 40.7060$_{\pm3.886}$      \\
+CoCoD(w/o PM)          & 0.8601$_{\pm0.003}$           & 0.7542$_{\pm0.005}$          & 0.7721$_{\pm0.002}$       & 33.1450$_{\pm1.058}$      \\
+CoCoD(w/ PM)            & $0.8779_{\pm0.001}$          & $0.7823_{\pm0.001}$            & 0.7811$_{\pm0.002}$       & 35.1829$_{\pm7.295}$     \\
+Graph Flow(w/o PM)            & $0.8816_{\pm0.002}$          & $0.7882_{\pm0.004}$            & 0.7832$_{\pm0.001}$       & 30.2157$_{\pm2.093}$     \\
+Graph Flow(w/ PM)                  & \textbf{0.8869$_{\pm0.001}$}          & \textbf{0.7967$_{\pm0.001}$}           & \textbf{0.7870$_{\pm0.002}$}       & \textbf{29.0594$_{\pm0.599}$}      \\ \hline

\end{tabular}
}
\end{center}
\vspace{-3mm}
\end{table}

\begin{figure}[ht]
\begin{center}
\includegraphics[width=0.8\linewidth]{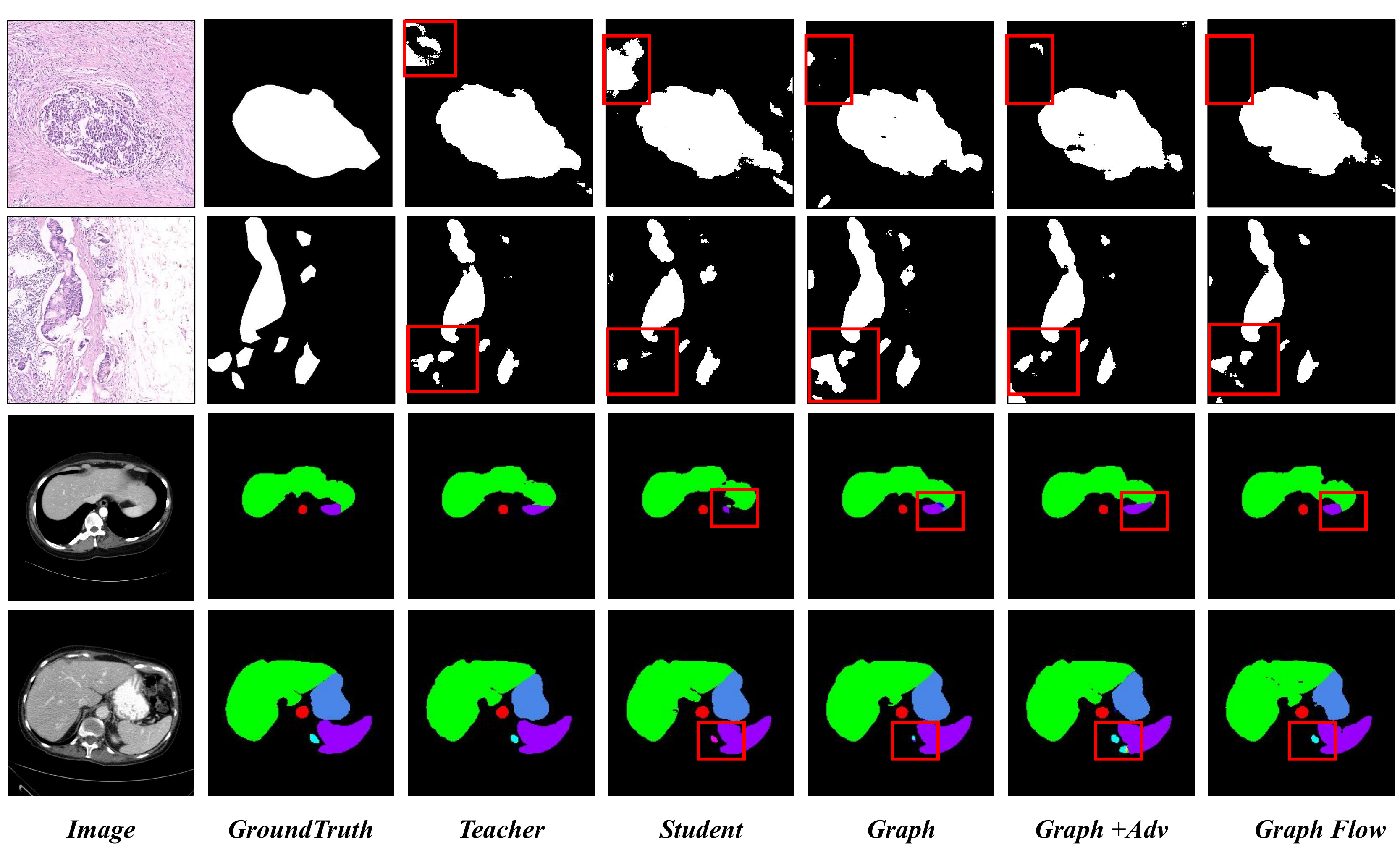}
\end{center}
   \vspace{-3mm}
   \caption{The visualization of components ablation on Gastric Cancer and Synapse. \textbf{Zoom in} for better details.}
   \vspace{-3mm}
   \label{fig:ablation_visualization}
\end{figure}

\textbf{\emph{2) Results of the Components Ablation:}} 
We conduct experiments to verify the contributions of each component in our Graph Flow. 
To begin with, we demonstrate the positive effectiveness of our three different distillation methods as shown in Table.~\ref{tab:components}.  
To be specific, our Graph Flow Distillation can induce $4.98\%$ improvement of  average DSC on Synapse, and Adversarial distillation can promote $0.44\%$ in mIOU for gastric cancer segmentation. 
Our overall Graph Flow can achieves a competitive performances compared with $T1$ ($90.05\%$ ACC vs. $88.69\%$ ACC on Gastric Cancer, $79.50\%$ average DSC 
vs. $78.70\%$ average DSC on Synapse).
In addition, we visualize the effectiveness of each component of our Graph Flow in
Fig.~\ref{fig:ablation_visualization}. 
%We could observe that our Graph Flow Distillation has more correct semantic class-wise knowledge than student trained from scratch (last third column). 
%Moreover, Adversarial distillation can complete the holistic alignment of the predictions between teacher and student (second row, last second column), while Logits Distillation is more beneficial for details (last column).

Afterwards, the indispensability of our Paraphraser Module is verified in Fig.~\ref{fig:loss_compare} and Table.~\ref{tab:Paraphraser}.
%As shown in Fig.~\ref{fig:loss_compare}, the Paraphraser Module can significantly stabilize training of our Graph Flow Distillation as it refines the knowledge of teacher by mitigating the gap between teacher and student. 
%Therefore, our Paraphraser Module can give $0.53\%$ promotion in ACC on Gastric Cancer and a $0.38\%$ in average DSC on Synapse. 
Our Paraphraser Module can give $0.53\%$ promotion on ACC for Gastric Cancer and $0.38\%$ on average DSC for Synapse. 
Besides, we can observe that the performance degradation of our Graph Flow is more slight compared with our previous work when the Paraphraser Module is subtracted.
%Therefore, our proposed Graph Flow is more robust than our previous work.}
%Furthermore, the reduction of HD from $35.1829$ mm to $29.0594$ mm on Synapse indicates our novel Graph Flow Distillation is more adapted than our CoCo DitillNet for Paraphraser Module.
%\vspace{-1mm}
%\vspace{-1mm}

\begin{figure}[ht]
\begin{center}
\includegraphics[width=0.8\linewidth]{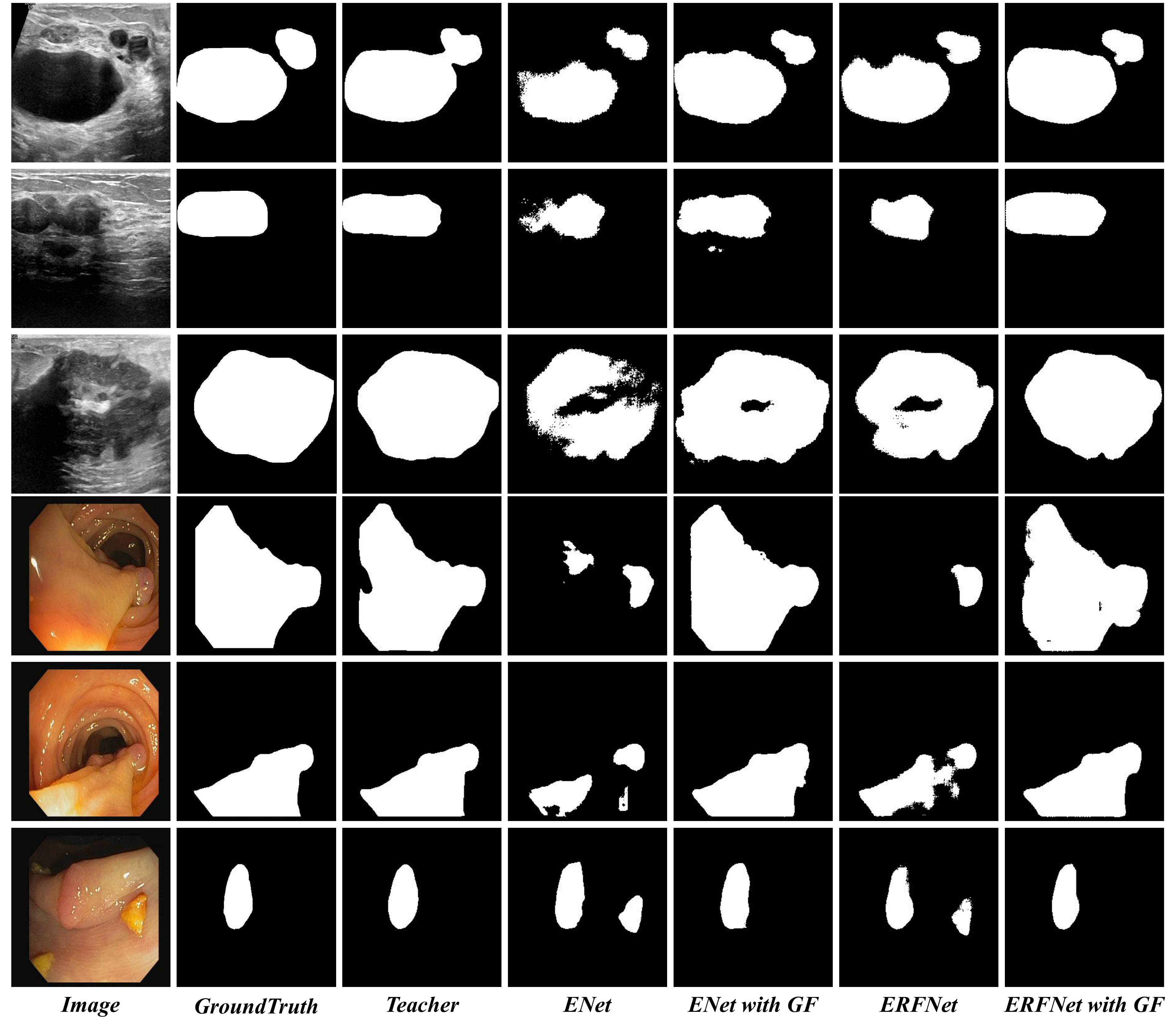}
\end{center}
   \vspace{-3mm}
   \caption{The visualization of generalization ablation on BUSI and CVC-ClinicDB. Teacher is TransUnet.
   GF indicates Graph Flow. 
   \textbf{Zoom in} for better details.}
   \vspace{-3mm}
   \label{fig:student_visualization}
\end{figure}

\textbf{\emph{3) Results of the Generalization Ablation:}} As shown in Table.~\ref{tab:student}, we leverage our Graph Flow on two different cumbersome networks: FANet and TransUnet, and three different lightweight networks: Mobile U-Net, ENet, and ERFNet.
It is obvious that these lightweight networks have lower complexity than cumbersome network, while their lightweight designs are disadvantage for performance.
However, our proposed knowledge distillation method can greatly improve the performance of such networks without any extra parameters. 
Especially, ENet, the most lightweight network, obtains only $1.56\%$ lower than FANet in ACC for gastric cancer pathology image segmentation.
Moreover, on CVC-ClinicDB, our method can improve the perfomance of ENet by $3.92\%$ in average DSC and $3.41\%$ in mIOU, when teacher networks is Transformer architecture.
In addition, the visualization results of different student networks (ENet and ERFNet) with our method on BUSI and CVC-ClinicDB, which are taught by TransUnet, are shown in Fig.~\ref{fig:student_visualization}.
%It indicates that diverse lightweight networks with our Graph Flow can completely substitute cumbersome networks in resource-limited clinical diagnosis.

%\vspace{-3mm}

\vspace{-3mm}
\subsection{Results of Semi-supervised Learning}
%Our method can improve the performance of compact networks without extra annotations compared with training from scratch.
%It could provide a novel semi-supervised paradigm by distilling knowledge from cumbersome networks.
%Therefore, We explore the ability of our method in \textbf{label-limited} situations at last.
At last, we explore the ability of our method in \textbf{label-limited} situations.
For Gastric Cancer, we randomly select labeled data in the whole datasets according to several different labeled data proportion ($20\%$, $40\%$, $60\%$, $80\%$, $100\%$).
The Synapse is collected from CT scans from different patients. 
A CT scan of a certain patient could be included in both labeled and unlabeled set, when we randomly select labeled data from the whole dataset according to these different labeled slices proportions.
It would lead to the leak of the information of the unlabeled set.
Therefore, for Synapse, we select a certain number of CT scans of patients as labeled data ($2$, $6$, $10$, $14$, $18$) and the rest of CT scans as unlabeled data.
Then, We conduct $4$ group of experiments in each fixed proportion or numbers of labeled data : 
$1)$ training a teacher network only with labeled data;
$2)$ training a student network only with labeled data;
$3)$ training a student network only with labeled data by our Graph Flow (pre-trained teacher network from $1)$) ;
$4)$ training a student network with labeled and unlabeled data by our Graph Flow (pre-trained teacher network from $1)$). 

\begin{figure}[ht]
\begin{center}
\includegraphics[width=0.9\linewidth]{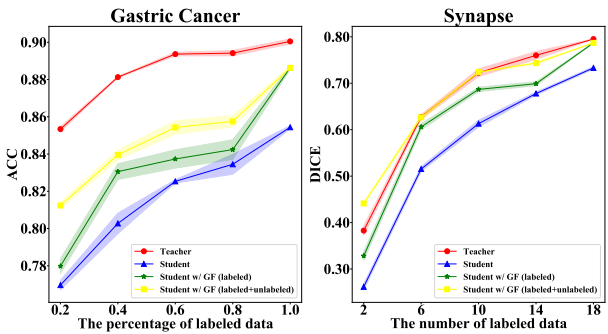}
\end{center}
   \vspace{-3mm}
   \caption{The results of Semi-supervised Learning with Graph Flow on Gastric Cancer and Synapse. 
   The bold line represents the metric means with five-time experiments. 
   The shadow regions indicate the fluctuations.
   The red line represents the performance of teacher. 
   The blue line represents the student trained from scratch.
   The green line represents the student trained with only labeled data. 
   The yellow line represents the student semi-supervised trained with all data. 
   GF indicates Graph Flow.}
   \vspace{-0mm}
   \label{fig:semi_supervisied}
\end{figure}

\begin{figure}[ht]
\begin{center}
\includegraphics[width=\linewidth]{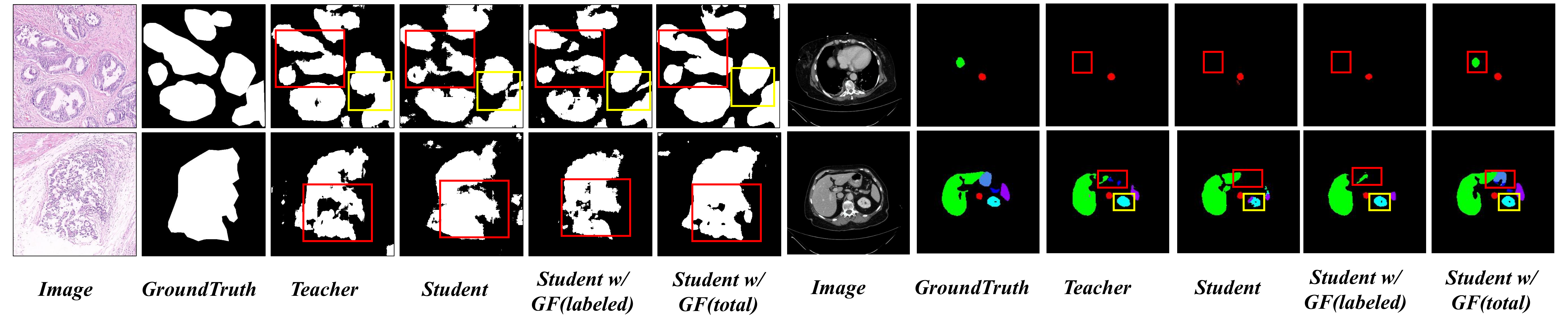}
\end{center}
   \vspace{-3mm}
   \caption{The visualized exhibition of semi-supervised learning on Gastric Cancer (the proportion of labeled data is $0.2$) (left) and Synapse (the number of labeled CT scans is $2$) (right). \textbf{Zoom in} for better details.}
   \vspace{0mm}
   \label{fig:semi_supervised_visualization}
\end{figure}

The results of semi-supervised learning on Gastric Cancer and Synapse are shown in Fig.~\ref{fig:semi_supervisied}. 
We also present the effectiveness visualization of semi-supervised learning in Fig.~\ref{fig:semi_supervised_visualization}. 
We can observe that the student network with labeled and unlabeled data in our GF framework outperforms the student network only with labeled data, and even surpass the teacher network (the number of labeled CT scans is $2$ on Synapse). 
In addition, the superiority of our method with all dataset (labeled + unlabeled) is more noteworthy, when the proportion or numbers of labeled data is more limited (the gap between the yellow and green curve tends to expand with the reduction of labeled data). 
%All of these confirm our method with unlabeled data can yield extra gains. 
%It indicates our method can simultaneously solve \textbf{resource-limited} and \textbf{annotation-limited} issues, which is so called \textbf{dual efficient} (network efficient and annotation efficient) medical image segmentation.
\section{Discussions and Limitations}
In this section, we give detailed discussions about the experimental results and the specific application scenarios of our method.
Furthermore, we discuss the method limitations.

\vspace{-3mm}
\subsection{Discussion on Different Experiments}
\textbf{\emph{1) Comparisons with Other Knowledge Distillation Methods:}} From the Table.~\ref{tab:kd}, we can observe that our method has statistically significant improvement compared with most state-of-the-art knowledge distillation methods.
Specifically, our method has remarkable reduction of HD on Synapse.
The lower HD means that the shape of our predictions are closer with the annotations.
This indicates that our method can better represent holistic semantic-aware knowledge, which is crucial for medical image segmentation.  
%From the perspective of generalization and robustness of the algorithms, we find that the KD methods have limitations at the different modalities, as reported in Table.~\ref{tab:kd}. 
Most of these methods in Table.~\ref{tab:kd} can only work well in one of the datasets.
On the contrary, our Graph Flow exhibits prominent performance on the both datasets with different image modalities obtained from pathological slides and CT scans, respectively.
We conjecture that these methods do not notice the knowledge hidden in the flow of cross-layers of teacher networks, which can teach student networks to extract more critical features for accurate organic or pathological segmentation.
Moreover, the variation of semantic-aware knowledge between different layers is general for medical image segmentation networks. 
It indicates that our method has better generalization ability on different modalities.

\textbf{\emph{2) Comparisons with Our Prior Work:}} From the last two rows of Table.~\ref{tab:kd} and Fig.~\ref{fig:visualization}, we can observe that our predictions are shape-closer with the annotations, as shown in the last row in Fig.~\ref{fig:visualization}.
Besides, our method could eliminate extra error predictions better than our previous work as shown in Fig.~\ref{fig:visualization} left.
Moreover, we conjecture that our method can teach student to capture the spatial relations between different categories, because our salience graphs can encode class-wise spatial information.
Moreover, our method also transfers the flow of inter-class correlation from teacher to student.
Our Graph Flow can better cooperate with the supervision of ground truth than our prior CoCo DistillNet.
On the other hand, our new method avoids the computing of Channel Mixed Spatial Similarity Module (CMSS) proposed by our previous work, which involves the resolution-wise multiplication operations ($B \times HW\times HW$) with the huge storage memory.
Therefore, our Graph Flow does not add complexity and needs less storage memory than our previous work with a better performance.

\textbf{\emph{3) Patch-wise Salience Region Ablation:}} The different patch-wise salience regions mean that the salience graph has different sizes of respective fields on class-wise knowledge. 
As shown in Table.~\ref{tab:patchsize}, the comparisons between local and non-local ($64\times64$) salience regions demonstrate that our salience operation can eliminate redundant knowledge, thus the salience graph could focus more on favorable semantic class-aware knowledge.
From Table.~\ref{tab:patchsize}, we can conclude that the dataset tends to select small patch-wise salience regions of $P$, if there are great differences in category size (e.g. Synapse). 
On the other hand, the dataset with resemblance in category size (e.g. Gastric Cancer) may employ relative large $P$. 

\textbf{\emph{4) Components Ablation:}} From Table.~\ref{tab:components} and Fig.~\ref{fig:ablation_visualization}, we could conclude that our Graph Flow predicts more correct semantic class-wise knowledge than student trained from scratch (the third column from last). 
Moreover, Adversarial distillation can complete the holistic alignment of the predictions between teacher and student (second row, penultimate column), while Logits Distillation is more beneficial for details (last column). As shown in Fig.~\ref{fig:loss_compare}, the Paraphraser Module can significantly stabilize training of our Graph Flow Distillation as it refines the knowledge of teacher by mitigating the gap between teacher and student. Besides, as shown in Table.~\ref{tab:Paraphraser}, the comparisons between Graph Flow and our previous work indicate that our proposed Graph Flow is more robust than our previous work. Furthermore, Graph Flow Distillation is more adapted with Paraphraser Module than our CoCo DitillNet.

\textbf{\emph{5) Generalization Ablation:}} The generalization ability for different networks of knowledge distillation methods is crucial in medical image segmentation.
The results of Table.~\ref{tab:student} demonstrate that our method can extract knowledge from different cumbersome networks (conventional convolutional architecture or prevalent transformer architecture), and then transfer the knowledge to different lightweight networks.
Furthermore, diverse lightweight networks with our Graph Flow can completely substitute cumbersome networks in resource-limited clinical diagnosis. 

\textbf{\emph{6) Semi-supervised Learning:}}All of these shown in Fig.~\ref{fig:semi_supervisied} and Fig.\ref{fig:semi_supervised_visualization} confirm our method with unlabeled data can yield extra gains. 
It indicates our method can simultaneously solve \textbf{resource-limited} and \textbf{annotation-limited} issues, which is so called \textbf{dual efficient} (network efficient and annotation efficient) medical image segmentation.
\vspace{-3mm}
\subsection{Discussion on Application Scenarios}
Medical image segmentation has achieved greatly progresses in the past several years, which benefits from the development of deep learning.
However, these networks with outstanding performances always need huge computational cost and inference time.
Therefore, they are impractical in some poverty-stricken rural hospitals of developing countries, especially in village or personal clinics. 
In these regions, extra purchasing of expensive computer-assisted diagnosis devices with GPUs may represent a heavy burden.
Furthermore, too long inference time of these networks could damage the efficiency of real-time computer-assisted diagnosis (e.g. intraoperative stent segmentation\cite{zhou2020lightweight}) for large hospitals in developed countries with tens of thousands of patients every day.
Meanwhile, the lack of large-scale annotated medical datasets further hinders applications of high-performance networks in the real world.

Our Graph Flow can balance the trade-off between the performance and consumption. 
Our method can improve lightweight networks with comparative performances than cumbersome networks, without extra computation cost and inference time.
Moreover, our method can achieve extra yields from extensive unlabeled medical images with a semi-supervised paradigm.
Therefore, to achieve medial image segmentation in the real world, our proposed Graph Flow could be employed.
\vspace{-3mm}
\subsection{Limitations} 
In this paper, we design a novel distillation method (i.e. Graph Flow Distillation) and propose a full-stage knowledge distillation framework combined with Adversarial Distillation and Logits Distillation. 
The distillation framework transfers the flow of semantic-aware knowledge, distills the knowledge of outputs, and utilizes the adversarial training on the prediction jointly.
Our Graph Flow exhibits superior performance on different medical modalities, different cumbersome networks, and different lightweight networks.
However, there are still some limitations in our method.
Firstly, the performance gap between the student trained by our method and the teacher still exists, which is also a general limitation of knowledge distillation. 
Though in some cases, the student trained by our method can surpass the teacher when the annotated images are scarce, 
%(the experiments of Synapse with 2 annotated patients CT images illustrated in Fig.\ref{fig:semi_supervisied}), 
further research is needed to explore the lightweight student network with the teacher level performance.
Secondly, our salience operation utilizes a fixed size of respective field for each channel.
The fixed respective field may lead to important area loosing, since we tend to utilize small respective field when the medical images have great category-size differences.
Though the results of patch-wise salience region ablation indicate that the category with large size may be insensitive for the size of respective field.
%In other words, the fixed respective field could not be damaged to the performance of our method.
If we could design more flexible respective sizes for different channels, our method will improve the student performance with better details.
At last, the semi-supervised learning schema and the knowledge distillation framework are embedded jointly.
From the perspective of semi-supervised learning, our semi-supervised strategy may be deficient compared with the methods specifically designed for tackling the lack of annotated data.
%semi-supervised learning. tackling the lack of annotated data
In addition, it is worth noting that our method is more applicable to harsh medical scenarios with data efficiency and network efficiency.

\section{Conclusion}
In this paper, we propose a knowledge distillation method, Graph Flow, based on the feature flow of salience features from teacher network and student network. 
With the distillation of cross-layer graph flow knowledge and the outputs, the Graph Flow can significantly improve the performance of a lightweight network for dual efficient medical image segmentation framework. 
On four different-modality and multi-category datasets, we conduct a series of experiments to demonstrate the effectiveness of our method.
Furthermore, our method provides a new paradigm for semi-supervised efficient medical image segmentation, which could jointly solve the scarceness of large-scale annotations and the resource-limited medical application problem. 
%In the future, we would like to focus on knowledge transfer from different modalities images for efficient medical image segmentation. 

\end{document}